\relax

\documentclass{svproc}

\usepackage{url}

\usepackage{times}
\usepackage{helvet}
\usepackage{courier}
\usepackage{graphicx}
\usepackage[square,sort,comma,numbers]{natbib}
\usepackage{caption}
\DeclareCaptionStyle{ruled}{labelfont=normalfont,labelsep=colon,strut=off}
\usepackage{subcaption}
\usepackage{multirow}
\begin{document}
\mainmatter            
\title{Laplacian Pyramid-Like Autoencoder}
\titlerunning{Laplacian Pyramid-Like Autoencoder}

\author{Sangjun Han\inst{1} \and Taeil Hur\inst{2}\textsuperscript{$\dagger$} \and Youngmi Hur\inst{3}\textsuperscript{$\ddagger$}}

\authorrunning{S. Han, T. Hur and Y. Hur}

\tocauthor{Sangjun Han, Taeil Hur, Youngmi Hur}
\institute{School of Mathematics \& Computing (Mathematics), Yonsei University, Seoul, South Korea,\\
\email{qkqhwl4@yonsei.ac.kr}
\and
JENTI Inc., Seoul, South Korea,\\
\email{taeil.hur@jenti.ai}
\and
Department of Mathematics, Yonsei University, Seoul, South Korea,\\
\email{yhur@yonsei.ac.kr}}

\maketitle

\begin{abstract}
In this paper, we develop the Laplacian pyramid-like autoencoder (LPAE) by adding the Laplacian pyramid (LP) concept widely used to analyze images in Signal Processing. LPAE decomposes an image into the approximation image and the detail image in the encoder part and then tries to reconstruct the original image in the decoder part using the two components. We use LPAE for experiments on classifications and super-resolution areas. Using the detail image and the smaller-sized approximation image as inputs of a classification network, our LPAE makes the model lighter. Moreover, we show that the performance of the connected classification networks has remained substantially high. In a super-resolution area, we show that the decoder part gets a high-quality reconstruction image by setting to resemble the structure of LP. Consequently, LPAE improves the original results by combining the decoder part of the autoencoder and the super-resolution network.
\end{abstract}

\section{Introduction}
\label{S:intro}
Deep neural networks are standard machine learning methods for diverse image processing such as object classification, image transform, image recognition. The networks have great varieties in architectures, algorithms, and processes. The autoencoder is a part of these varieties. 

The autoencoder encodes a given data to some representation in a latent space, usually compressed from the input data, by a few layers. Then it decodes this representation to the reconstruction converted to have desired properties by different layers. The encoder has the advantage of analyzing the data in a low dimensional space, in a way similar to the Principal Component Analysis. Also, the simplicity of the model structure makes it easy to modify the structure according to various purposes like unsupervised pre-training, denoising, restoration of image. 

Our paper is motivated by the article \cite{WAE}. The authors develop the wavelet-like autoencoder (WAE) and use it for acceleration in classification networks. WAE decomposes the input image into two down-scaled images, low-frequency information $I\textsubscript{L}$ and high-frequency information $I\textsubscript{H}$, through the encoder and reconstructs the original image at the decoder. Their use of the prefix ``wavelet-like'' is due to the imposed condition on $I\textsubscript{H}$ to be sparse and the reconstruction process obtained by adding the convolution filtered versions of $I\textsubscript{L}$, $I\textsubscript{H}$. In WAE, to accelerate the classification model, they input $I\textsubscript{L}$ as the mainstream and $I\textsubscript{H}$ as a helper to the classification networks (e.g., VGG16, ResNet50) instead of the original image. The change made the network have smaller computational complexity; hence it takes less time for the entire process. Besides, the complementary analysis using the helper $I\textsubscript{H}$ makes the network stay competitive in terms of accuracy. 

Although WAE is good at accelerating the basic classification networks, it is not satisfactory in some crucial aspects. First, contrary to the wavelet, WAE does not impose any condition on low-frequency information. This missing condition makes the approximation $I\textsubscript{L}$ hard to reflect the original image, which can drop classification performance. Second, such a low-frequency image can lower the quality of the reconstruction image. The WAE paper does not pay attention to the reconstructed result because its primary concern is the acceleration problem, requiring only two decomposed images in the classification networks. This limited architecture makes it hard for use in other areas requiring a reconstruction. Third, the name of the autoencoder is ``wavelet-like,'' but the model is missing a critical feature of the wavelet, namely multi-scale property. Consequently, WAE has difficulty decomposing multiple times, resulting in a restriction on WAE so that the model is stuck in the acceleration task.

Considering the preceding, we propose a new model named the Laplacian pyramid-like autoencoder (LPAE). We impose an extra condition on the low-frequency part of WAE, and get an autoencoder with a hierarchy, similar to the shape of the Laplacian pyramid (LP) introduced in \cite{LP}. As a result, LPAE makes the approximation image with better quality. Using this approximation image, we obtain higher performance of classification but also extend to super-resolution problems with $2^k$ magnification for various $k$ since LPAE decomposes and reconstructs an image multiple times.

The datasets used for the classification are ImageNet2012 (ImageNet) from \cite{imagenet} and Intel Image Classification (Natural Scene) from  \cite{natural}. We combine two base networks, VGG16 (VGG) and ResNet50 (ResNet), with WAE and LPAE. LPAE shows better classification performance than WAE. LPAE accelerates test times sufficiently, although there is a slight drop in the acceleration time. In some cases, LPAE even has a faster total training time than WAE. For a super-resolution problem, we use three datasets, CelebA \cite{celeba}, DIV2K \cite{div2k}, and Set5 \cite{set5}. After training on CelebA, the test is done on CelebA. After training on DIV2K, the test is carried out on Set5, following the convention in the field. For the network, base network is WaveletSRNet (WaveSR) introduced in \cite{WSR}. Since WaveSR uses the wavelet packet transform for image reconstruction, we can directly see the effectiveness of LPAE for super-resolution by replacing the wavelet part with LPAE. This result in super-resolution shows the potential usefulness of the proposed LPAE for solving other problems in deep learning because the classification and the super-resolution are different types of problems.

\section{Related Works}
\subsection{Autoencoder}
Recently, most research about the autoencoder concentrates on its connection to other applications rather than developing it independently. Generative autoencoder \cite{AAE,TVA} occupies a significant portion as a base model for application. In \cite{crash}, variational autoencoder (VAE) solves a problem of insufficient data by producing synthetic data from a given sample data. \cite{VAESR} develops a reference-based SR approach using  VAE, which can take any image as a reference. \cite{NVAE} suggests a novel generative model with hierarchical structure based on VAE. By exploiting unusual convolutions such as depthwise separable convolution and regular convolution, authors get high performance without modifying the loss function of VAE. Other existing autoencoders, such as sparse autoencoder and denoising autoencoder, are used to solve specific problems. For example, \cite{electricity} composes stacked sparse denoising autoencoder for detecting electricity theft and \cite{landslide} exploits a sparse autoencoder for landslide susceptibility prediction. Because the autoencoder extracts the feature efficiently and reconstructs data with a simple structure, it is a popular tool for developing a framework combined with machine learning. 
Our model, LPAE, is a model possessing the properties of LP. So LPAE can provide new approaches to diverse problems where LP properties are helpful.

\subsection{Network Acceleration}
After the substantial progress of the convolutional neural network (CNN), a lot of research focuses on the acceleration of the networks with keeping high performance. There are several approaches to accelerate networks. \cite{WAE, architecture} propose methods that modify an architecture or overall structure. \cite{architecture} suggests an accelerator using the point that can compute the convolution layer similar to the fully connected layer. Moreover, to accelerate model training, some researches present a new training framework. \cite{autoassist} makes an assistant model along with the main model to remove trivial instances during training, then gets the results with the new training algorithm. \cite{autofreeze} suggests a new training plan for fine-tuning. It accelerates the fine-tuning by adaptively freezing the nearly converged portion of the network. Since many natural language processing and classification models exploit fine-tuning the pre-trained network, this method is tempting. Furthermore, the study to achieve harmony between the hardware and the software is actively performed \cite{tensordash, hardsoft}. Considering rapidly advanced hardware, both propose a hardware-based approach to speed up a CNN and reduce the energy required for computation.

\subsection{Single Image Super-Resolution}
In the super-resolution area, many of the studies try to build a deep convolution-based model using up-sampling. For example, \cite{VDSR} introduces the way of the deep network for super-resolution by a residual learning and high initial learning rate, while \cite{EDSR} suggests enhancing the power of the deep network by removing the batch normalization and setting a training pipeline. \cite{DBPN} makes a model to repeat the up-scaling and down-scaling iteratively and to reflect the feedback of error. In addition to the construction of deep models, some researchers are focusing on other points. By learning the feature correlations of hidden layers by second-order attention, \cite{secondattention} improves the expressional power of CNN. \cite{CAR} trains the content-adaptive resampling (CAR) model with the main super-resolution (SR) network to create a low-resolution image. Since unsupervised training is conducted on CAR, CAR makes a down-scaled image to keep important information for super-resolution impartially. The primary concern of \cite{accurate, classSR} is the speed of the SR. While both of them construct a new network structure and framework, neural architecture search is a noticeable characteristic of \cite{accurate}. The LPAE that we propose in this paper is an assistant model to reconstruct images trained with the main SR model, similar to the approach in \cite{CAR}. Also, LPAE learns the correlation between the components for super-resolution and can improve the reconstructing power of the main SR model similar to \cite{secondattention}. However, LPAE is an autoencoder with a straightforward structure and is easy to connect with diverse architectures without too much modification.

\section{Laplacian Pyramid in Neural Network}

\subsection{Idea and Structure of LP}
The Laplacian pyramid (LP) is introduced in \cite{LP} as a technique for compact image encoding. This technique has its root in subtracting a low-pass filtered image from the original image. Such subtraction reduces redundant information by decreasing the correlation between neighboring pixels in an image, in other words, data compression. 
Moreover, this compression process can be repeated on low-pass filtered images having different scales. Then the repetitions form a pyramid-like structure and accelerate the reduction. Since this process is similar to the Laplacian operator sampling on diverse scales, the pyramid is named the Laplacian pyramid.

The following is the overall process of establishing LP.
For an input image $I_{0}$, a filtered image $I_{1}$ is obtained with the decreased resolution by some low-pass filter. A filtered image $I_{2}$ is similarly obtained from $I_{1}$, and proceeding successively results in a sequence $\{I_{k}\}_{k=0}^{K}$.
For each fixed $I_{k}$ with $k\ge 1$, the image $\tilde{I}_{k}$ with the same size as $I_{k-1}$ is defined by expanding $I_k$ by interpolation. Subtracting $\tilde{I}_k$ from $I_{k-1}$ produces $d_k$ with the same size as $I_{k-1}$ but with compressed image information. The sequence $\{d_{k}\}_{k=1}^{K}$ of such differences corresponds to the pyramid of LP.

\subsection{Strengths of LP and Applications on Neural Net}
According to the above procedure, LP decomposes an image $I$ into a low-pass filtered image $I_{c}$ and a difference $I_{d}$ after one stage. The decomposition result of LP is redundant compared to the wavelet transform often used as the analysis tool because the $I_{d}$ part of LP has the same size as the original image. But some advantages are originated from the structure of LP. For example, the implementation that requires only low-pass filtering and subtraction is straightforward. Another strength of LP is that there exists no scrambled frequency \cite{Framing} derived from high-pass filtering. And the organizational style of LP enables the perfect reconstruction of the original image. Such advantages encourage many usages of LP as a tool in traditional image processing.

In addition, there have been lots of efforts in machine learning to treat LP, both independently and as a concept in a model. For the super-resolution problem, \cite{DRLN} and \cite{LPcgal} get competitive results by introducing the style of LP in the attention module and by considering a generator of the conditional generative adversarial network, respectively. On the other hand, \cite{transfer, photoreal} use LP itself to make input images. After generating the image pyramid from the original image, \cite{transfer} put components into two modules for high-quality transfer, a drafting module that transforms artistic style from low-resolution image and a revision module that refines the transformed image with a high-resolution image. On top of these, there are applications for diverse problems such as object detection \cite{LFPN}, image compression \cite{compression}. These researches show a possibility of a connection between LP and deep learning, presenting satisfactory results. As will be seen later in this paper, our LPAE presents convincing results, which add to the above possibility.

\section{Laplacian Pyramid-like Autoencoder}
\label{S:LPAE}

We propose a simple autoencoder model, the Laplacian pyramid-like autoencoder (LPAE), to have properties of LP such as a hierarchical structure, analysis and reconstruction parts. Then LPAE is connected to the classification networks and the super-resolution networks. These connection is performed to show the effectiveness of LPAE.

\begin{figure}[h]
\centering
\includegraphics[width=0.5\textwidth]{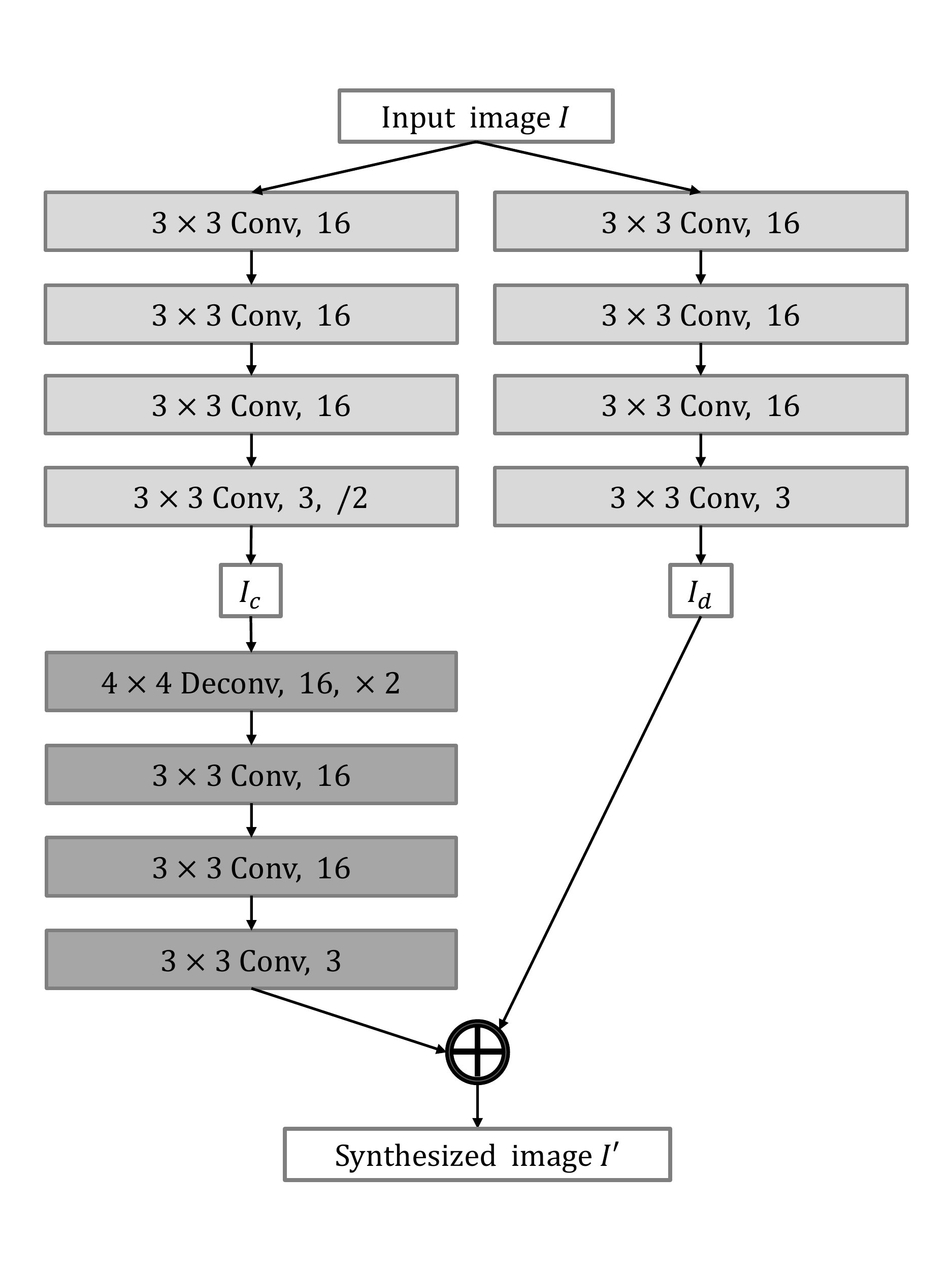}
\caption{Overall structure of LPAE.}
\label{fig1}
\end{figure}

\subsection{Proposed Model}
LPAE has two parts, encoding (analysis) and decoding (reconstruction) parts. Fig.~\ref{fig1} shows the overall structure of LPAE. In the analysis part, LPAE decomposes the original image $I$ into images $I_c$ and $I_d$. $I_c$ is the down-sampled approximation image that is low-pass filtered, and $I_d$ is the detail image representing the difference between the original image and the prediction from approximation. Both outputs pass through $4$ convolution layers. When we downsample images, we use the convolution layer with stride $2$ instead of a pooling layer. In the reconstruction part, the output image $I'$ is reconstructed by an element-wise sum between the detail image $I_d$ and the prediction image $\phi(I_c)$, where $\phi$ is a deconvolution process. The process $\phi$ consists of up-sampling with $4\times4$ transposed convolution layer and filtering with $3$ of convolution layers. 
In our LPAE, except for the output layers, we use convolution layers with filter-size $3\times3$ and 16 output channels for simplicity and efficiency, but the network can be made deeper or more redundant.

\begin{figure}
\centering
\begin{subfigure}[b]{0.48\textwidth}
\includegraphics[width=0.96\textwidth]{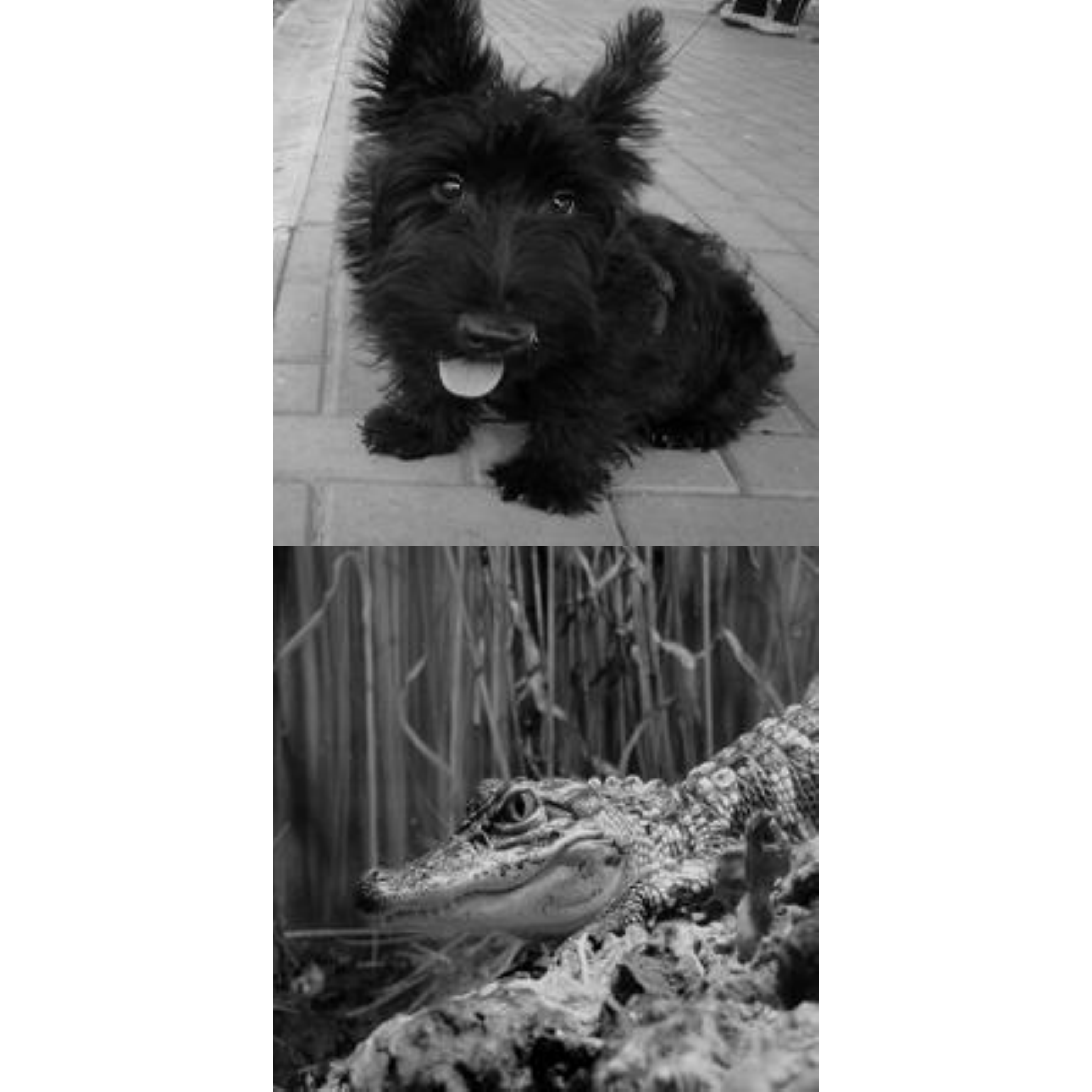} 
\captionsetup{justification=centering}
\caption{Original image \\ \phantom{x}}
\end{subfigure} 
\begin{subfigure}[b]{0.48\textwidth}
\includegraphics[width=0.96\textwidth]{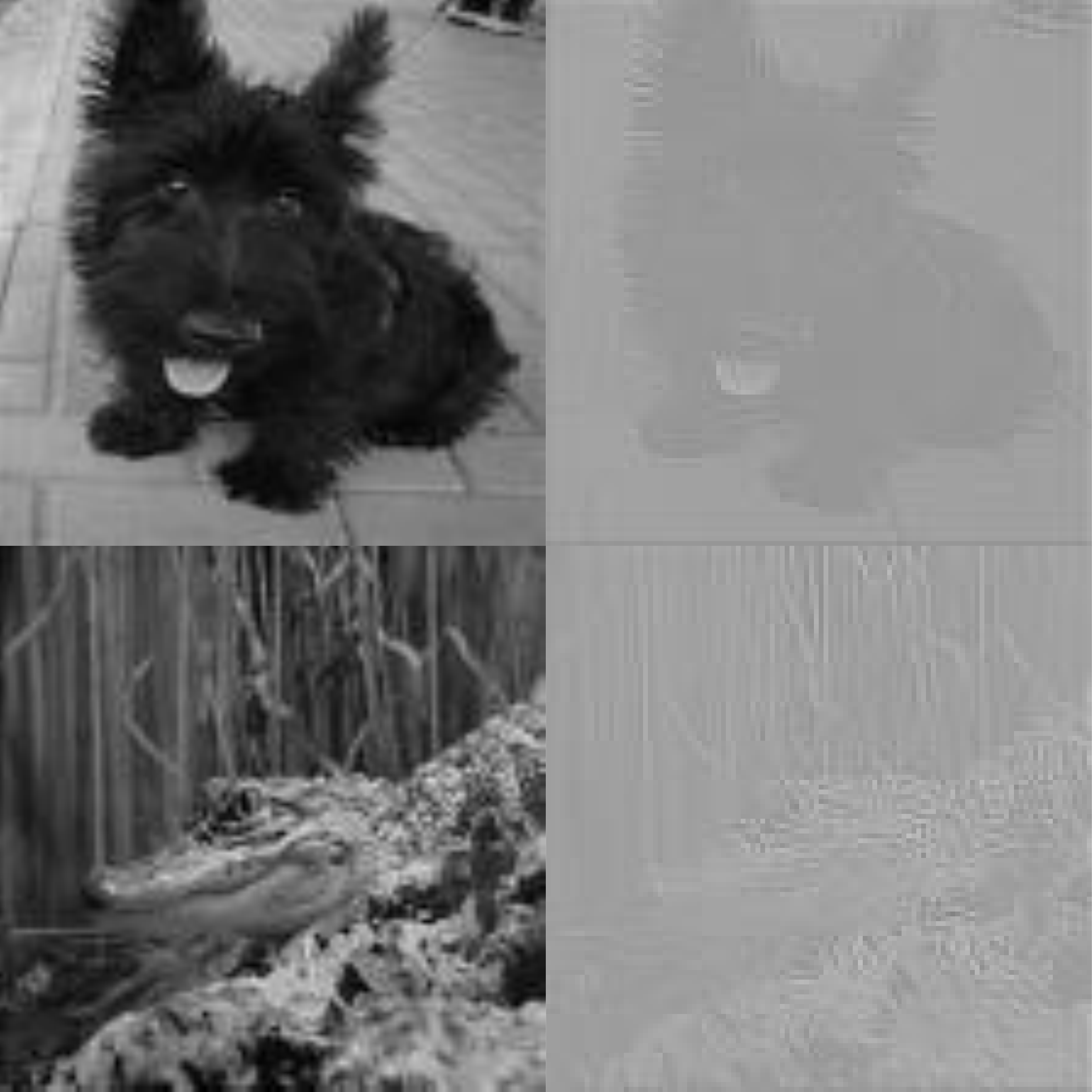} 
\captionsetup{justification=centering}
\caption{Approximation image \\ (Left: LPAE, Right: WAE)}
\end{subfigure}
\begin{subfigure}[b]{0.48\textwidth}
\includegraphics[width=0.96\textwidth]{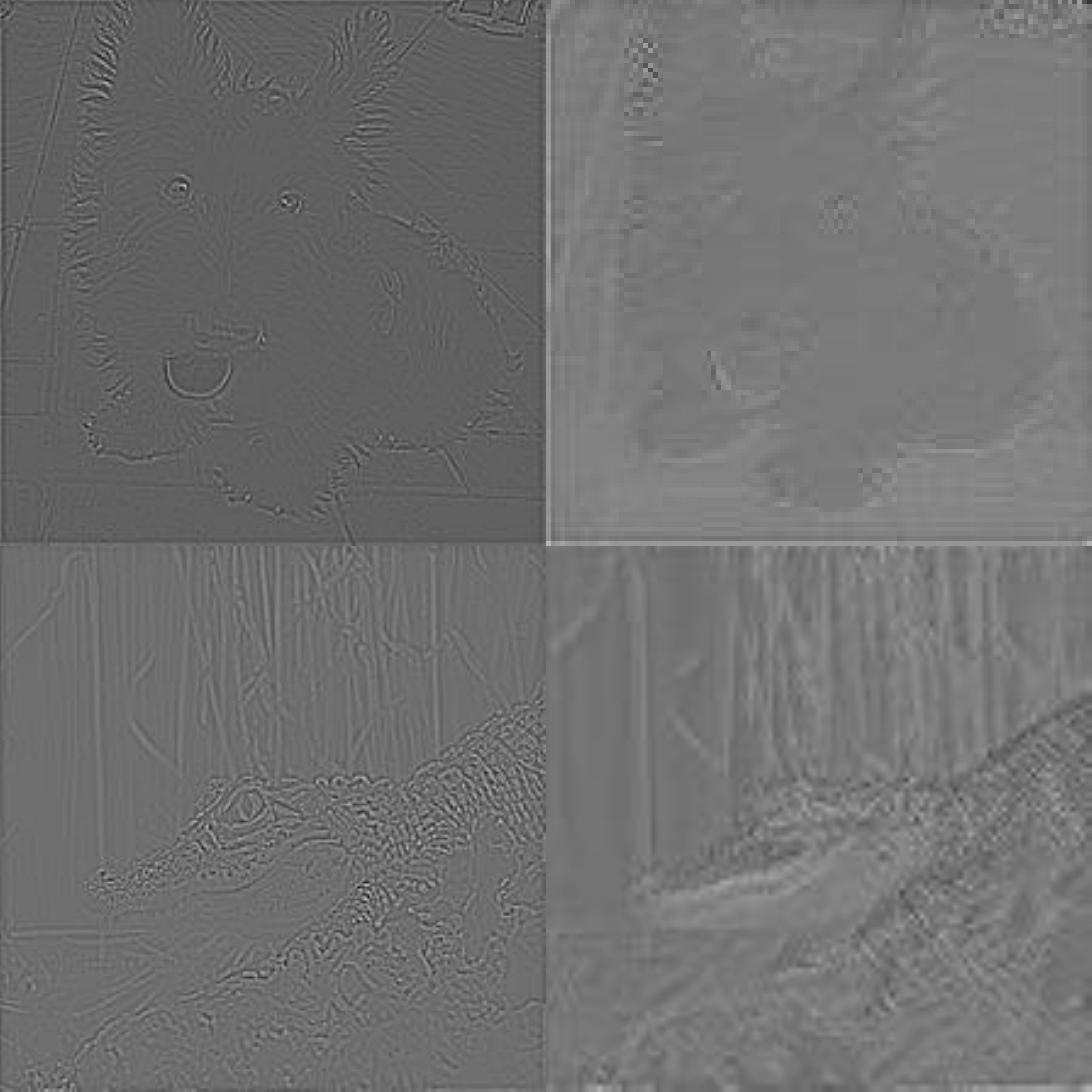} 
\captionsetup{justification=centering}
\caption{Detail image \\ (Left: LPAE, Right: WAE)}
\end{subfigure} 
\begin{subfigure}[b]{0.48\textwidth}
\includegraphics[width=0.96\textwidth]{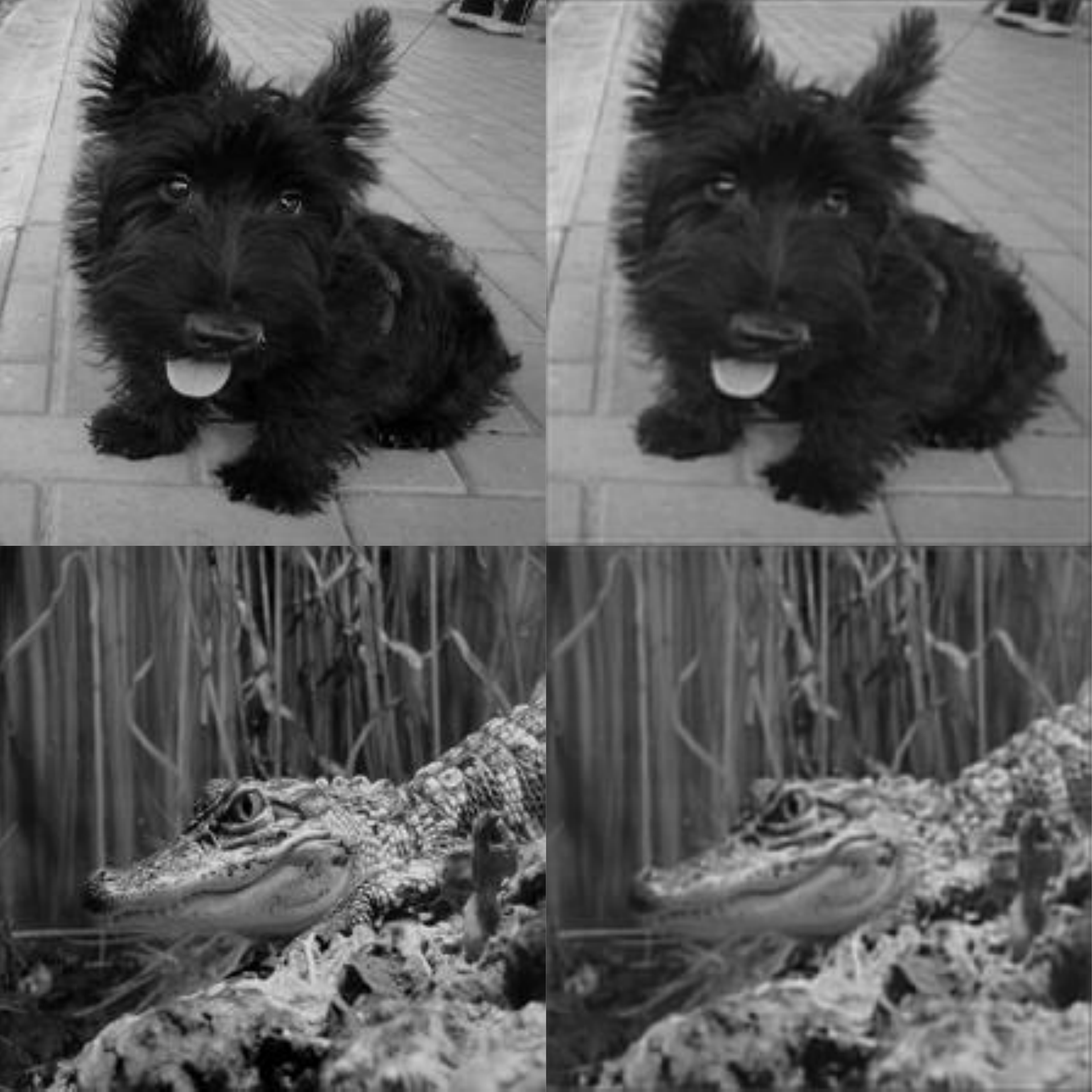} 
\captionsetup{justification=centering}
\caption{Reconstruction image \\ (Left: LPAE, Right: WAE)}
\end{subfigure}
\caption{Results of LPAE and WAE for two images from ImageNet. Original images are in (a). In (b)-(d), left images are LPAE results, and right images are WAE results.}
\label{fig2}
\end{figure}

\subsection{Loss for LPAE}
Our loss function consists of three components that make the autoencoder similar to LP. Since we want LPAE to reconstruct the original image as much as possible, we define our first loss function to be the reconstruction loss, $$l_r=\frac{1}{|I|}||I-I'||_1.$$

For the approximation image representing the low-frequency channel of the original image well, we prepare an approximation image $I_\downarrow$ by using the bicubic interpolation and apply the mean square error (MSE). we then define our second loss function as the energy (or approximation) loss, $$l_e=\frac{1}{|I_{c}|}||I_c-I_\downarrow||_{2}^{2}.$$
For the approximation image $I_\downarrow$, many other methods, including the wavelet transform by CDF 9/7 filters, can also be used, but we find no significant difference in the loss for using different approximation methods. Hence in this paper, we fix the bicubic interpolation to get $I_\downarrow$. The bicubic interpolation makes a natural connection with the super-resolution problem.

To constrain the detail image to be sparse, we set our last loss function to be the sparsity loss, 
$$l_s=\frac{1}{|I_{d}|}||I_d||_2^2.$$
This sparsity loss makes the detail image carry a high-frequency channel of the original image and provide textures to the connected network.

The overall loss of LPAE is defined as the weighted sum of the three losses: $$l_{total}=\alpha l_r+\beta l_e+\gamma l_s,$$
where $\alpha =\gamma =1, \beta =0.8$ to give less weight on the approximation loss than other losses.

Fig.~\ref{fig2} shows the result of LPAE and WAE applied to two different images from ImageNet. For the approximation image in Fig.~\ref{fig2}(b), there is no doubt that the result of LPAE is more vivid, original-like image. Besides, Fig.~\ref{fig2}(c), the result of detail images, demonstrates that this precise approximation influences the detail image to be sparse. Although sparse, the detail contains many information for the original texture since it has the same spatial size as the original image, unlike WAE. Eventually, LPAE gets a more sharp reconstruction, as seen from the images in Fig.~\ref{fig2}(d). The differences prove the validity of our loss function.

\subsection{Image Classification Problem}
In \cite{WAE}, the authors show that WAE can accelerate classification networks while keeping the accuracy almost the same by using two outputs of the encoding part. We show that LPAE can accelerate classification networks at about the similar level, even with a slight improvement in accuracy.
We think that using both the approximation and the detail, similar to \cite{WAE}, has contributed to getting comparable accuracy, even if the input in VGG is not the original image. In this sense, we speculate that there is an improvement in accuracy for LPAE because LPAE gets a better approximation and a high-resolution detail than WAE.

To be comparable with the experiments in \cite{WAE}, we set the same structure except for the autoencoder part, replacing WAE with LPAE. Below we describe the use of the autoencoder only for VGG, as the process for ResNet is similar. Recall that the encoder part of LPAE decomposes an original image $I$ into an approximation image $I_c$ and a detail image $I_d$. The features $f_c$ for $I_c$ are extracted from the feature extraction part of VGG. The features $f_d$ for $I_d$ are extracted from another lighter feature extraction part consisting of convolution layers with only a quarter of the output channels for $f_c$. Using fully connected layers, we obtain the classification score $s_c$ from $f_c$, and $s_d$ from the concatenation of $f_c$ and $f_d$. The final score $s$ is the average of the two scores $s_c$ and $s_d$.

From the acceleration perspective, according to \cite{WAE}, the total complexity of all the convolutional layers can be represented by
\begin{equation}
    O(N), \hbox{ where } N=\sum_{l=1}^{d}{n_{l-1}\cdot s_{l}^{2}\cdot n_{l}\cdot m_{l}^{2}}.
    \label{ComputaionalComplexity}
\end{equation}
Here, $d$ is the number of the convolution layers. For the $l$-th layer, $n_l$ is the number of the output channels, ${s_l}$ is the size of the kernels, and ${m_l \times m_l}$ is the spatial size of the output features.

Since LPAE has $\frac{1}{2}$ size of feature maps for the approximation, the complexity $N$ in (\ref{ComputaionalComplexity}) is $\frac{1}{4}$ of the original network.
However, unlike WAE, there is no change in the size of feature maps for the detail. Still following the setup of WAE for the detail, the number of the output channels for the LPAE's detail becomes $\frac{1}{4}$ compared to the approximation case, so the complexity becomes $\frac{1}{16}$ of the original network. As a result, LPAE has $\frac{5}{16}$ total complexity compared to the original. Thus the acceleration rate is about $1 / \frac{5}{16}=3.2$. This number is comparable to $3.76$, the acceleration rate of WAE.

\subsection{Applications to Super-Resolution Problem}
As observed earlier, LPAE's approximation image in Fig.~\ref{fig2} tends to be similar to some other low-pass filtered images, such as the approximation image of the wavelet transform or the bicubic interpolation. This approximation image helps the detail to be more sparse.
Besides, LPAE can make the hierarchical structure because the approximation carries low-frequency information sufficiently. Based on these observations, we try to expand the application domain of LPAE to super-resolution.

There are lots of algorithms in the super-resolution problem. Some of them are hard to be connected to LPAE directly in a natural way. So we try to choose the models having room for substitution, and WaveletSRNet (WaveSR) in \cite{WSR} is one such model. The basic concept of the network is the use of the wavelet packet transform (WPT) as a reconstructor. For example, to train WaveSR for magnification of $4$, an input image is decomposed into two levels using WPT. Then the network gets the approximation image of $\frac{1}{4}$ size and extracts features from the approximation image through the embedding part. The features form new details of the number needed in the reconstruction process (15 for magnification of 4) and one approximation image. At last, through WPT, the network creates the high-resolution image of magnification of 4. In this procedure, the authors set the loss function of WaveSR to make the new details and approximation similar to the decomposition results of WPT.

In this algorithm, we insert LPAE as a substitute for WPT. The divided encoding and decoding parts can take the role of the existing decomposition and reconstruction, respectively. Moreover, the hierarchical structure of LPAE does not restrict its use only for a magnification of 2 but allows it for an arbitrary magnification of $2^k$. Although LPAE produces larger-sized details than WPT, the redundancy of information is helpful to reconstruct. And the fact that it has a smaller number of detail images makes the network efficient. In addition, we expect the substitute to improve the high-resolution result because it has the flexibility that the reconstruction of LPAE gets better by a fine-tune on various datasets in contrast with WPT.

We modify the loss function given in \cite{WSR} to fit LPAE's situation when we train WaveSR substituted with LPAE (named LPSR). The reconstruction loss is defined by using the L1 distance between the reconstruction $I'$ and the original high-resolution image $I$,
$$l_{rec}=\frac{1}{|I|}||I-I'||_1.$$ 

The use of L1 loss for the whole procedure can reduce the smoothing effect and enhance the quality of the image \cite{loss}. Based on this observation, we define the pyramid loss by 
$$l_{p}=\frac{1}{|I_{c}|}||I_c-I'_c||_{1}+\sum_{i}{\lambda_{i}\frac{1}{|I_{d_{i}}|}||I_{d_{i}}-I'_{d_{i}}||_{1}}$$
where $\{\lambda_{i}\}$ are weights for the detail part. For instance, we take $\lambda_{1}=0.8, \lambda_{2}=1.2$ in case of magnification of 4.

Finally, the loss function of LPSR is the weighted sum of two losses:
$$l_{total}=\gamma l_{rec}+\delta l_{p}.$$
If the new details $I'_{d_{i}}$ and approximation $I'_c$ produced by LPSR are close to the outputs $\{I_{d_{i}}, I_c\}$ of LPAE sufficiently, the reconstruction using the new ones will have high quality on high-resolution stage. From this point of view, we focus on the closeness between the outputs of LPAE and LPSR. Hence in all of our experiments, weights $\gamma=1$ and $\delta=10$ are used.

\subsection{Details on Experiments}
In classification, as mentioned in our paper, we use two backbones, VGG16, ResNet50, and use two datasets, ImageNet2012 (ImageNet), Intel Image Classification datasets (Natural Scene). ImageNet is a large dataset having train images of 1.2 million approximately and validation images of 50k. The images are classified as labels of 1000 and are with varying image sizes. In comparison to ImageNet, Natural Scene is a tiny dataset of 6 categories. There are 14k in train, 10k in validation and test. Moreover, the spatial size of each image is fixed as $150\times150$.

We mostly follow the training strategies in \cite{WAE} to train the autoencoders for comparison with WAE. In particular, for both WAE and LPAE, the Xavier algorithm is used for initialization of parameters, and the SGD algorithm with momentum 0.9 and weight decay of 0.0005 is used. However, some options are chosen differently because of the dataset size. Although we keep a batch of 4 as in \cite{WAE} for Natural Scene, we choose 256 for ImageNet to shorten training time. Also, our training epochs are fixed as 100 for Natural Scene and 20 for ImageNet. Considering the difference between the two autoencoders, we choose the initial learning rate differently. For WAE, it is set as 0.000001 with the decay factor of 0.1 after every 10 epochs as in \cite{WAE}, but for LPAE as 0.01 with the same decaying strategy.

To train the connected classification network, for Natural Scene images, we randomly crop to $128\times128$, and for ImageNet images, we resize to $256\times256$ and randomly crop to $224\times224$. The only data augmentation we select is the random horizontal flip. We choose the batch size to be 256 regardless of datasets, and choose the SGD algorithm with the same options. But the decay strategy and the number of epochs are different along with the dataset. For ImageNet, training is performed with 20 epochs, and then the learning rate is multiplied by 0.1 after 10 epochs. For Natural Scene, the learning rate is multiplied by 0.5 after every 10 epochs during 100 training epochs.

In the super-resolution, we train on the DIV2K dataset and test on the Set5 dataset. Additionally, we use the CelebA dataset for comparison with the original task of WaveSR.
DIV2K has 800 diverse images of the large size of 2k resolution, but CelebA has 162,770 images of center-arranged (to some degree) face with the size of $178\times218$.
Because of this substantial dissimilarity, we perform different data augmentations on two datasets.

The autoencoders' training options are chosen similarly to the classification. For DIV2K, we randomly crop the high-resolution image to $192\times 192$ with random horizontal/vertical flips. We train LPAE using the Adam algorithm with a batch of $4$ and adjust the initial learning rate of $0.001$ to be divided by $2$ after every $50$ epochs during $400$ epochs. To train LPSR, we set a batch of $8$ and select the Adam algorithm with the initial learning rate of $0.001$ decaying to half after every 50 epochs for $300$ training epochs. 
For the other dataset, CelebA, the training image is resized to $144\times144$ then randomly cropped to $128\times128$ with a random horizontal flip only. LPSR is trained during $40$ epochs using the Adam algorithm with a batch of $256$, the initial learning rate of $0.01$ multiplied by $0.1$ after every $10$ epochs. 
Options for other networks, WaveSR and WSR, are chosen to be identical to LPSR.
All of our codes are available at https://github.com/sangjun7/LPAE.

\section{Experimental Results}
In this section, we examine our LPAE performances. All of the detailed options on experiments are reported in Supplementary Material. We first make a comparison between LPAE and WAE using the PSNR value in Table~\ref{t1}. Then we join two autoencoders to the classification networks, VGG and ResNet. These networks represent the efficiency and the power of LPAE by inference time and accuracy.
The following super-resolution results show the versatility of LPAE. By checking PSNR and SSIM values, we show that replacing original reconstruction parts with LPAE enhances super-resolution abilities. All of the experiments are conducted on 6 GPUs of GeForce RTX 3090 except for measuring test time. We measure the test time of classification networks by a GPU of Tesla T4 on Google Colab.

\subsection{Autoencoder}

LPAE is motivated by WAE. However, there is a considerable difference between the two autoencoders in their network structure and loss formulation. As we can see in Fig.~\ref{fig2}, LPAE gets a high-resolution detail image compared to the detail image of WAE. The large-sized detail image has more abundant information for textures and high-frequency data of the original image.
Furthermore, LPAE makes a more specific approximation image than WAE, as seen in Fig.~\ref{fig2}. This fact comes from the constraint on the approximation image to resemble the bicubic interpolated image. Also, the change of the MSE loss between the input image and the reconstructed image to the L1 loss with a large learning rate returns high-quality reconstruction \cite{loss}. Table~\ref{t1} shows a difference between the two autoencoders. The PSNR values in the table are calculated between the original image and the reconstruction image. Although PSNR alone cannot determine the quality of reconstruction, many researchers indeed consider a large PSNR value as an essential indication for better reconstruction. For Bicubic, we put back to the original image using bicubic interpolation after reducing the spatial size to half. For ImageNet, we get the large PSNR value of 47.89 dB, far superior to the others, i.e., 28.57 dB for WAE and 28.44 dB for Bicubic. In the DIV2K dataset, LPAE makes the closer reconstruction image to the original than the case of ImageNet. The PSNR value of LPAE is 54.73 dB, and that of WAE is 19.90 dB, which shows a significant difference between PSNR values. LPAE gets one compressed image and another sparse image, and based on the above results, we see that LPAE restores the original image much better. Thus LPAE can accomplish the role of accelerator comparable to WAE and be extended to the super-resolution problem.

\begin{table}[t]
\centering
\begin{tabular}{cc|cc|c}
\hline\hline
\noalign{\vskip 0.5mm}
\multicolumn{2}{c|}{} & LPAE & WAE & Bicubic \\
\noalign{\vskip 0.5mm}
\hline
\noalign{\vskip 0.5mm}
\multirow{3}{*}{ImageNet} & Train Loss & 0.0041 & 0.0019 & - \\
& Test Loss & 0.0042 & 0.0019 & - \\
& PSNR (dB) & 47.89 & 28.57 & 28.44 \\
\noalign{\vskip 0.5mm}
\hline
\noalign{\vskip 0.5mm}
\multirow{3}{*}{DIV2K} & Train Loss & 0.0023 & 0.0114 & - \\
& Test Loss & 0.0024 & 0.0109 & - \\
& PSNR (dB) & 54.73 & 19.90 & 26.19 \\
\noalign{\vskip 0.5mm}
\hline
\hline
\end{tabular}
\vskip 5mm
\caption{Comparison of power for restoring the original image after encoding and decoding between WAE, LPAE and bicubic interpolation (Bicubic).}
\label{t1}
\end{table}

\subsection{Classification}
\begin{table}
\centering
\begin{tabular}{cc|ccc}
\hline\hline
\noalign{\vskip 0.5mm}
\multicolumn{5}{c}{ImageNet} \\
\hline
\noalign{\vskip 0.5mm}
\multicolumn{2}{c|}{\multirow{2}{*}{}} & Top 5 & Train & Trainable \\
\multicolumn{2}{c|}{} & Accuracy (\%) & Time (hr) & Parameters \\
\noalign{\vskip 0.5mm}
\hline
\noalign{\vskip 0.5mm}
\multirow{3}{*}{{\rotatebox[origin=c]{90}{VGG}}} & Basic & 86.94 & 30.11 & 138,365,992 \\
 & WAE  & 84.96 & 27.48 & 132,914,278 \\
 & LPAE & 85.32 & 27.12 & 150,220,870 \\
 \noalign{\vskip 0.5mm}
\hline
\noalign{\vskip 0.5mm}
\multirow{3}{*}{{\rotatebox[origin=c]{90}{ResNet}}} & Basic & 84.01 & 29.23 & 25,575,784 \\
 & WAE  & 79.95 & 27.68 & 29,628,406 \\
 & LPAE & 80.31 & 27.69 & 29,633,494 \\
 \noalign{\vskip 0.5mm}
\hline\hline
\noalign{\vskip 0.5mm}
\multicolumn{5}{c}{Natural Scene} \\
\hline
\noalign{\vskip 0.5mm}
\multicolumn{2}{c|}{\multirow{2}{*}{}} & Top 5 & Train & Trainable \\
\multicolumn{2}{c|}{} & Accuracy (\%) & Time (hr) & Parameters \\
\noalign{\vskip 0.5mm}
\hline
\noalign{\vskip 0.5mm}
\multirow{3}{*}{{\rotatebox[origin=c]{90}{VGG}}} & Basic & 89.97 & 0.55 & 138,365,992 \\
 & WAE  & 88.73 & 0.35 & 132,914,278 \\
 & LPAE & 89.97 & 0.38 & 150,220,870 \\
 \noalign{\vskip 0.5mm}
\hline\hline
\end{tabular}
\vskip 5mm
\caption{Comparison of accuracy and training times between classification networks connected with WAE or LPAE.}
\label{t2}
\end{table}

As mentioned in Section~\ref{S:LPAE}, we expect that the link between LPAE and a classification network reduces the computational cost and accelerates algorithms with a slight drop in accuracy. To show this, we train the basic classification networks (VGG, ResNet) and those connected with WAE and LPAE, using two datasets (ImageNet, Natural Scene). We use our codes for WAE. Although not able to reproduce exactly, we identify a tendency of accelerating presented in \cite{WAE}. Table~\ref{t2} shows a comparison of performances between networks. For all cases, classification networks connected with LPAE (LPVGG, LPResNet) get better precision than those connected with WAE (WVGG, WResNet). For Natural Scene, resulting in 89.97$\%$, LPVGG even has the same accuracy as the original VGG. We speculate that more information of the original image is kept after the LPAE's encoding helps classify. Table~\ref{t2} also reports the number of trainable parameters for each case for reference. The training times of LPVGG/LPResNet are reduced to a similar level as WVGG/WResNet.  For ImageNet, WVGG saves about 2.63 hr than the basic, and LPVGG saves about 2.99 hr that is about 10$\%$ of the whole training time of the basic. For Natural Scene, LPVGG reduces about 31$\%$ of training time than the basic.

\begin{table}
\centering
\begin{tabular}{cc|ccccc}
\hline\hline
\noalign{\vskip 0.5mm}
\multicolumn{7}{c}{ImageNet} \\
\hline
\noalign{\vskip 0.5mm}
\multicolumn{2}{c|}{\multirow{2}{*}{}} & {FLOPs} & {Compl.} & \multicolumn{3}{c}{Test Time (ms) for Batch} \\
\multicolumn{2}{c|}{} & {(B)} & {(B)} & 1 & 20 & 50 \\
\noalign{\vskip 0.5mm}
\hline
\noalign{\vskip 0.5mm}
\multirow{3}{*}{{\rotatebox[origin=c]{90}{VGG}}} & B & 31.02 & 15.47 & 14.45 & 127.07 & 274.55 \\
 & W  & 8.95 & 4.46 & 10.84 & 57.17 & 117.25 \\
 & L & 11.01 & 5.48 & 12.67 & 80.77 & 171.98 \\
 \noalign{\vskip 0.5mm}
\hline
\noalign{\vskip 0.5mm}
\multirow{3}{*}{{\rotatebox[origin=c]{90}{ResNet}}} & B & 7.77 & 4.06 & 7.40 &  61.09 & 140.10\\
 & W & 2.74 & 1.40 & 8.75 & 40.86  & 88.27\\
 & L & 3.68 & 1.88 & 9.26 & 56.90  & 127.46\\
 \noalign{\vskip 0.5mm}
\hline\hline
\noalign{\vskip 0.5mm}
\multicolumn{7}{c}{Natural Scene} \\
\hline
\noalign{\vskip 0.5mm}
\multicolumn{2}{c|}{\multirow{2}{*}{}} & {FLOPs} & {Compl.} & \multicolumn{3}{c}{Test Time (ms) for Batch} \\
\multicolumn{2}{c|}{} & {(B)} & {(B)} & 1 & 20 & 50 \\
\noalign{\vskip 0.5mm}
\hline
\noalign{\vskip 0.5mm}
\multirow{3}{*}{{\rotatebox[origin=c]{90}{VGG}}} & B & 10.15 & 5.06 & 6.96 & 41.78 & 85.88 \\
 & W  & 2.95 & 1.47 & 7.03 & 21.03 & 39.45 \\
 & L & 3.62 & 1.80 & 7.56 & 27.80 & 57.68 \\
 \noalign{\vskip 0.5mm}
\hline\hline
\end{tabular}
\vskip 5mm
\caption{Comparison about FLOPs and test times between VGG or ResNet connected with WAE and LPAE. B, W and L is Basic, WAE and LPAE same as Table~\ref{t2}, respectively. The unit of FLOPs and Compl. is a billion (B).}
\label{t3}
\end{table}

Table~\ref{t3} shows each network's FLOPs, complexity (Compl.), and test times for different batch sizes. Our complexity here is calculated similarly to $N$ in (\ref{ComputaionalComplexity}) for convolution layers and fully connected layers. For instance, with VGG on ImageNet, we get an acceleration rate of $\frac{15.47}{5.48} \approx 2.82$ for LPAE, and $\frac{15.47}{4.46} \approx 3.47$ for WAE. These numbers are similar to the rough computation of acceleration rate in Section~\ref{S:LPAE}, i.e., $3.2$ for LPAE and $3.76$ for WAE.
When we check the results of FLOPs about VGG, we get 31.02 billion (B) for the basic, 8.95 B for LPVGG, and 11.01 B for WVGG. Thus the FLOPs of LPVGG and WVGG are about 0.35 and 0.29 of that of VGG. Hence, the computational cost of LPVGG, similar to WVGG, is vastly reduced. We think that this point leads to the result accelerating LPVGG sufficiently in the test. LPVGG gets 12.67 ms in the test with 1 batch, which is decreased by 1.78 ms. If we raise batch size to 20 or 50 (cf. \cite{divbatch}), the decrease rate becomes larger because LPVGG has 80.77 ms instead of 127.07 ms for size 20, and 171.98 ms instead of 274.55 ms for size 50. Although the connection with LPAE accelerates VGG less than the connection with WAE, there is a meaningful difference in test time between LPVGG and VGG. For other cases with 1~batch, results are unexpected, representing that basic VGG takes the shortest time for the test. We think the unexpected situations are due to the small size of the input. For increased batch, the test time of networks connected with LPAE again becomes smaller than that of original networks.

\subsection{Super-resolution}

\begin{table}
\centering
\begin{tabular}{ccc|ccc}
\hline\hline
\noalign{\vskip 0.5mm}
 & Scale &  & WaveSR & WSR & LPSR \\
\noalign{\vskip 0.5mm}
\hline
\noalign{\vskip 0.5mm}
\multirow{6}{*}{{\rotatebox[origin=c]{90}{CelebA}}} & \multirow{2}{*}{$\times{2}$} & PSNR (dB) & 30.87 & 31.93 & 36.04 \\
& & SSIM & 0.920 & 0.940 & 0.967 \\
& \multirow{2}{*}{$\times{4}$} & PSNR (dB) & 27.71 & 17.96 & 29.15 \\
& & SSIM & 0.840 & 0.513 & 0.871 \\
& \multirow{2}{*}{$\times{8}$} & PSNR (dB) & 24.34 & 15.05 & 24.86 \\
& & SSIM & 0.707 & 0.431 & 0.735 \\
\noalign{\vskip 0.5mm}
\hline
\noalign{\vskip 0.5mm}
\multirow{6}{*}{{\rotatebox[origin=c]{90}{Set5}}} & \multirow{2}{*}{$\times{2}$} & PSNR (dB) & 32.03 & 25.70 & 37.62 \\
& & SSIM & 0.914 & 0.781 & 0.955 \\
& \multirow{2}{*}{$\times{4}$} & PSNR (dB) & 28.60 & 15.28 & 32.25 \\
& & SSIM & 0.852 & 0.381 & 0.899 \\
& \multirow{2}{*}{$\times{8}$} & PSNR (dB) & 24.01 & 13.85 & 27.07 \\
& & SSIM & 0.650 & 0.336 & 0.782 \\
\noalign{\vskip 0.5mm}
\hline\hline
\end{tabular}
\vskip 5mm
\caption{Results for the super-resolution network based on WaveletSRNet.}
\label{t4}
\end{table}

\begin{table}
\centering
\begin{tabular}{c|ccc}
\hline\hline
\noalign{\vskip 0.5mm}
 {\multirow{2}{*}{}} & \multicolumn{3}{c}{Set5} \\
  {} & $\times{2}$ & $\times{4}$ & $\times{8}$ \\
\noalign{\vskip 0.5mm}
\hline
\noalign{\vskip 0.5mm}
CAR \cite{CAR} & 38.94 & 33.88 &  \\
DRLN+ \cite{DRLN} & 38.34 & 32.74 & 27.46 \\
ABPN \cite{ABPN} &  & 32.69 & 27.25 \\
HBPN \cite{HBPN} & 38.13 & 32.55 & 27.17 \\
DBPN-RES-MR64-3 \cite{DBPN} & 38.08 & 32.65 & 27.51 \\
MWCNN \cite{MWCNN} & 37.91 & 32.12 &  \\
CARN \cite{CARN} & 37.76 & 32.13 &  \\
LFFN-S \cite{LFFN} & 37.66 & 31.79 &  \\
CSRCNN \cite{CSRCNN} & 37.45 & 31.01 & 25.74 \\
IKC \cite{IKC} & 36.62 & 31.52 &  \\
DeepRED \cite{DeepRED} &  & 30.72 & 26.04 \\
\noalign{\vskip 0.5mm}
\hline
\noalign{\vskip 0.5mm}
LPSR & 37.62 & 32.25 & 27.07 \\
\noalign{\vskip 0.5mm}
\hline\hline
\end{tabular}
\vskip 5mm
\caption{Comparison of PSNR values of LPSR with the State-of-the-art (SOTA) on Set5.}
\label{t5}
\end{table}

\begin{figure}
\centering
\begin{subfigure}[b]{0.19\textwidth}
\includegraphics[width=0.95\textwidth]{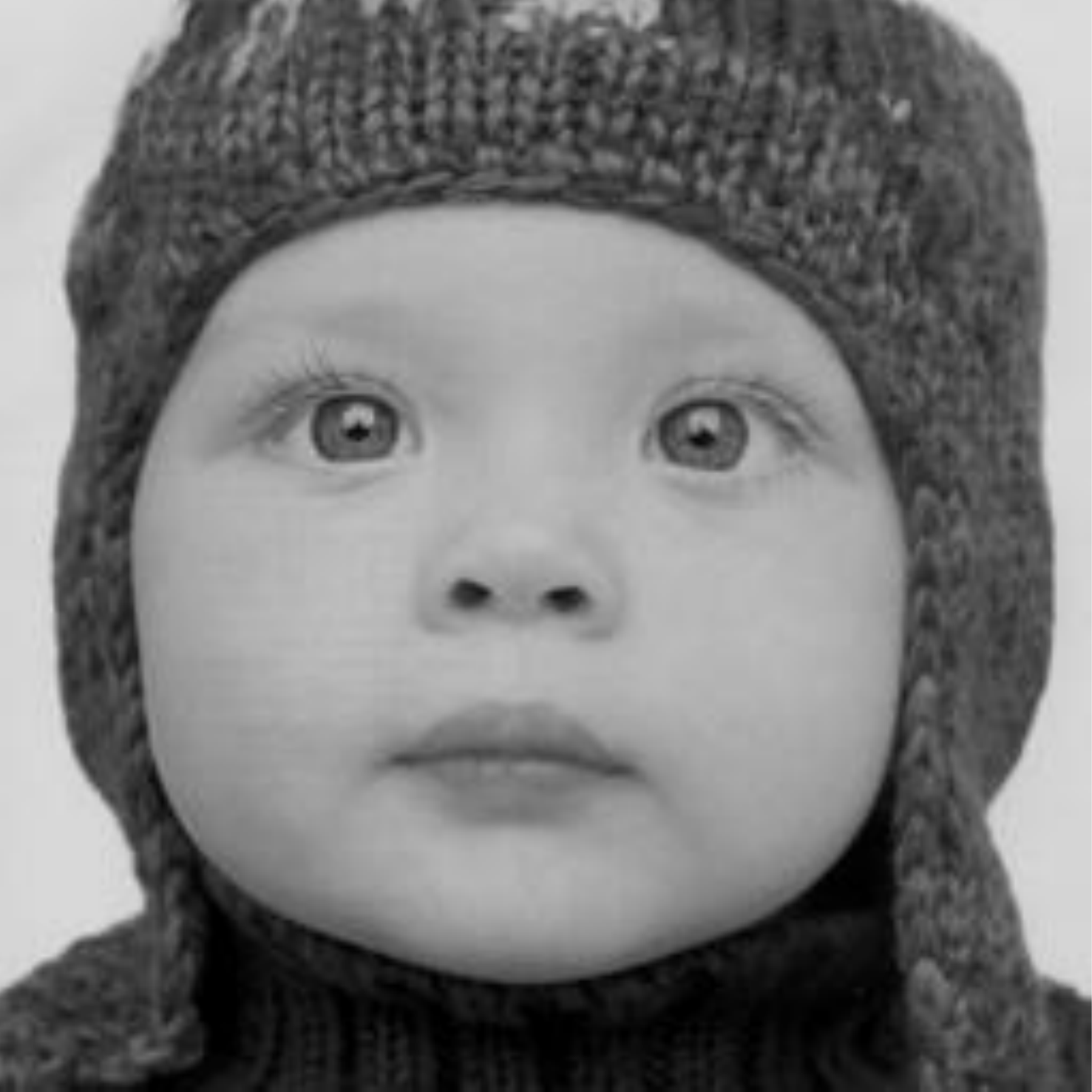} 
\end{subfigure}
\begin{subfigure}[b]{0.19\textwidth}
\includegraphics[width=0.95\textwidth]{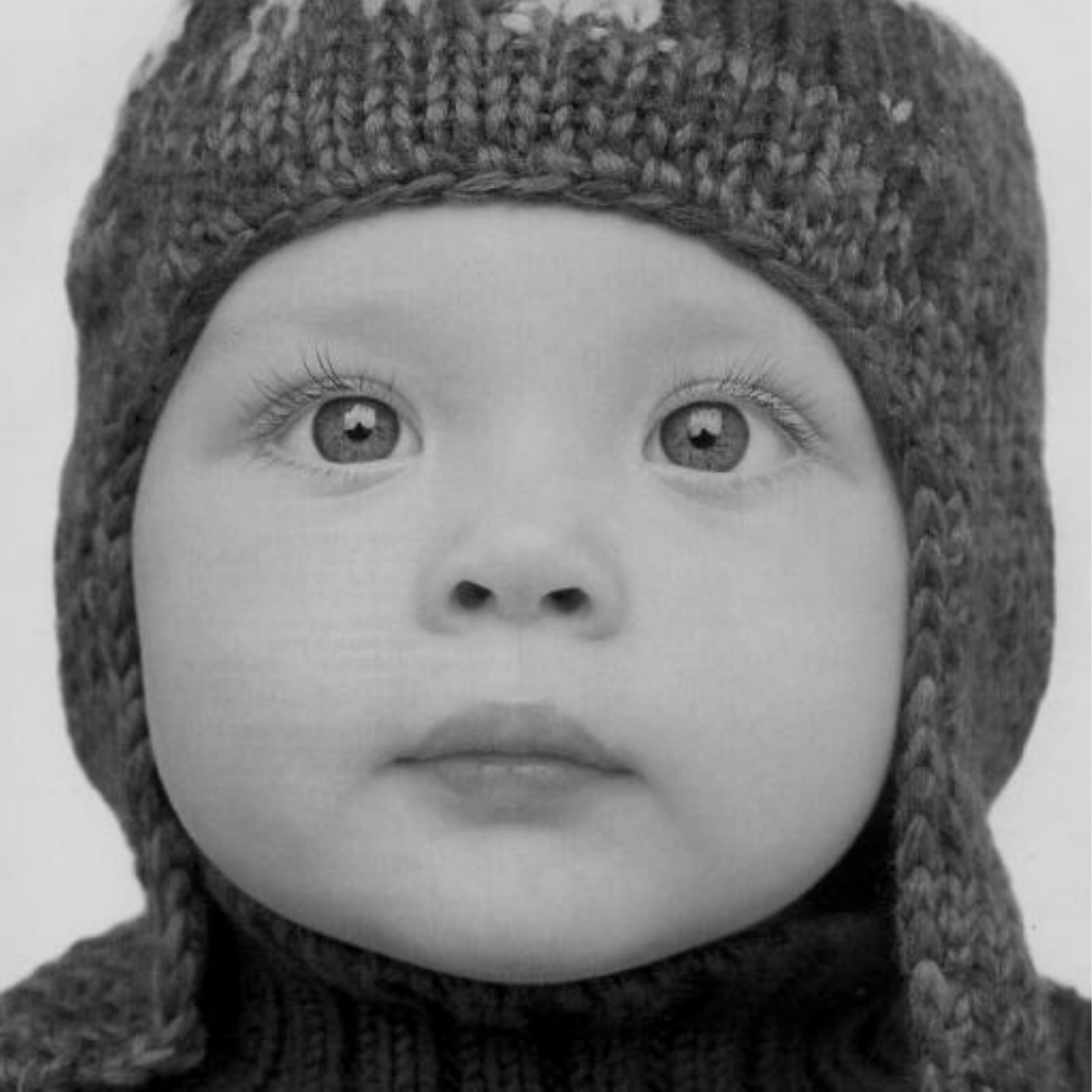} 
\end{subfigure}
\begin{subfigure}[b]{0.19\textwidth}
\includegraphics[width=0.95\textwidth]{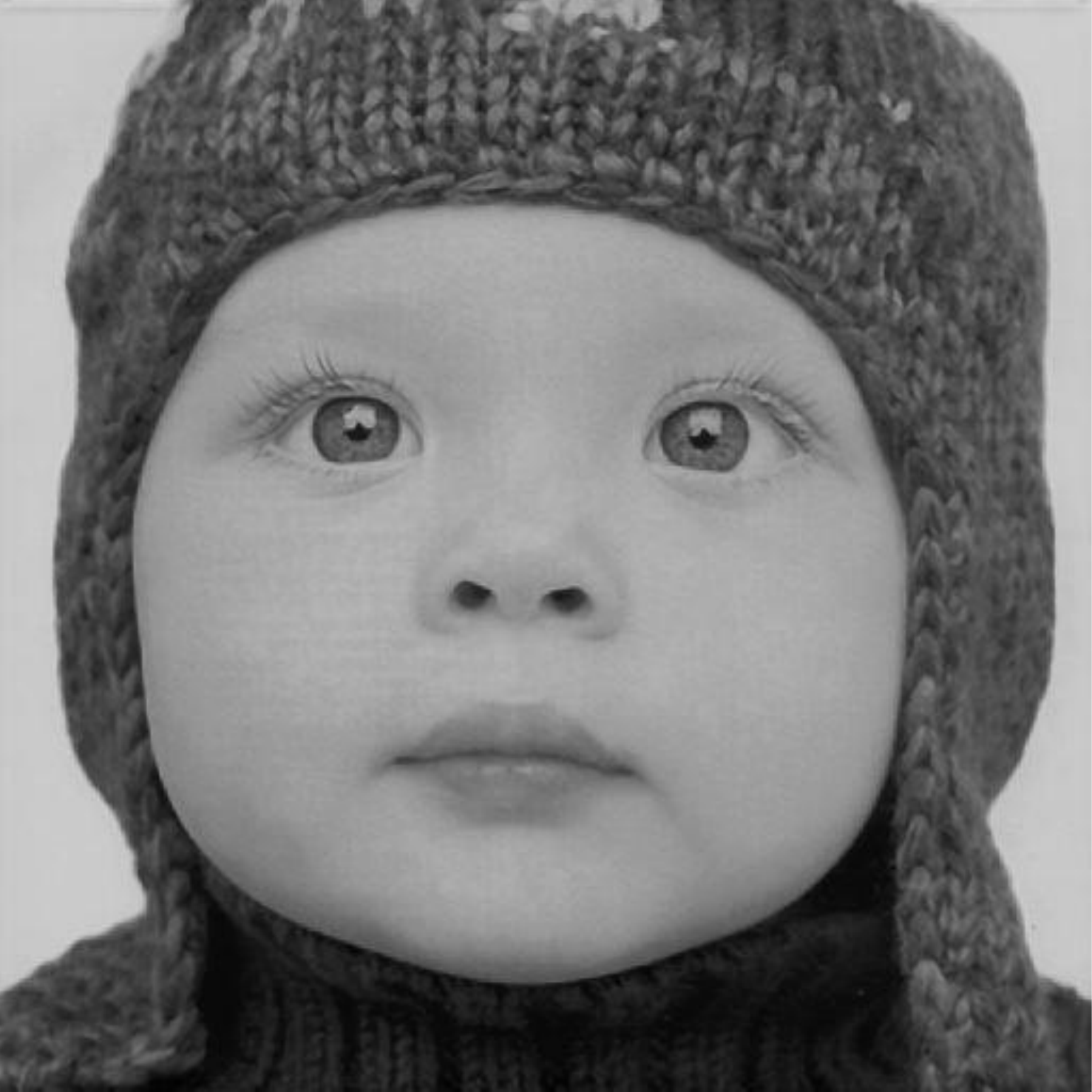} 
\end{subfigure}
\begin{subfigure}[b]{0.19\textwidth}
\includegraphics[width=0.95\textwidth]{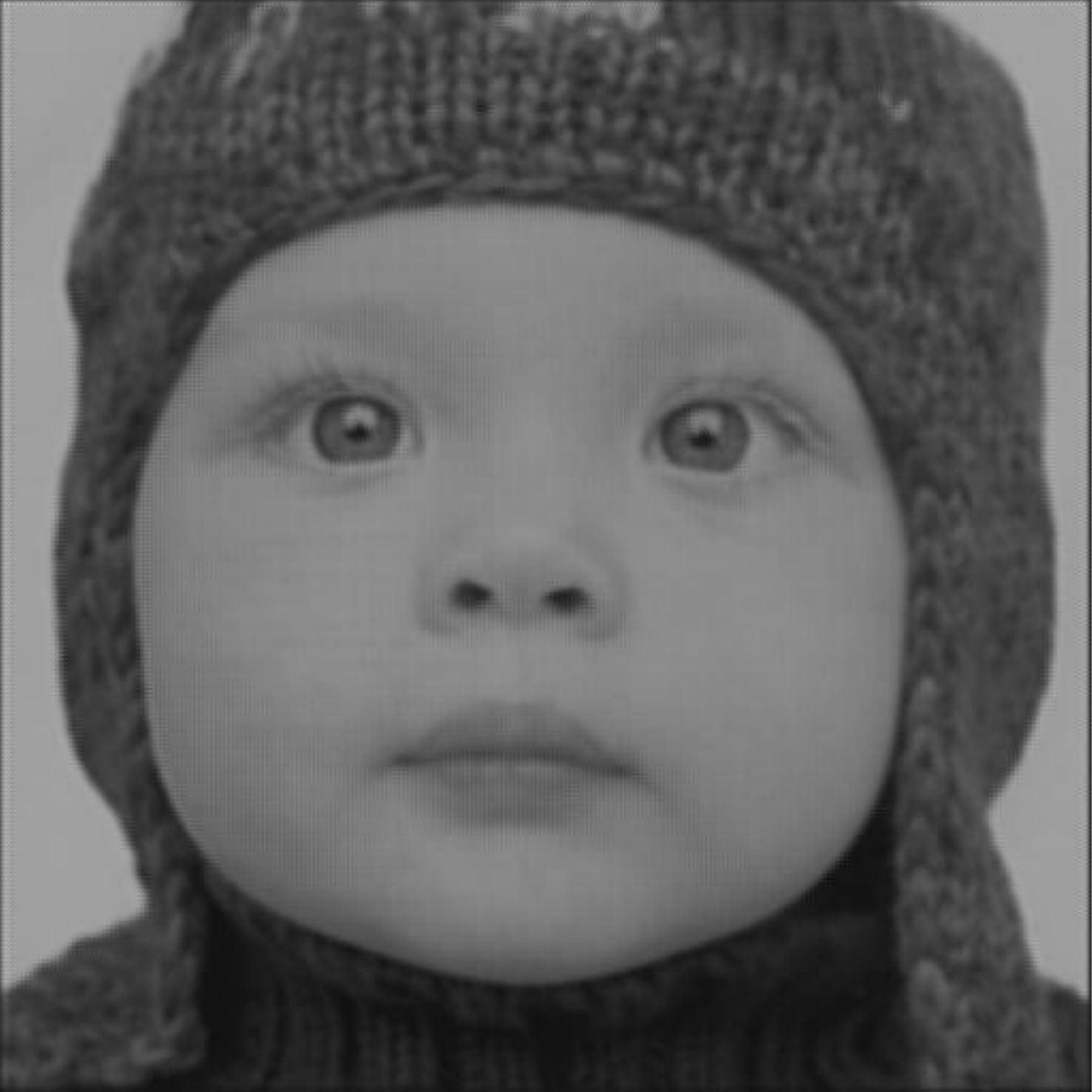} 
\end{subfigure} 
\begin{subfigure}[b]{0.19\textwidth}
\includegraphics[width=0.95\textwidth]{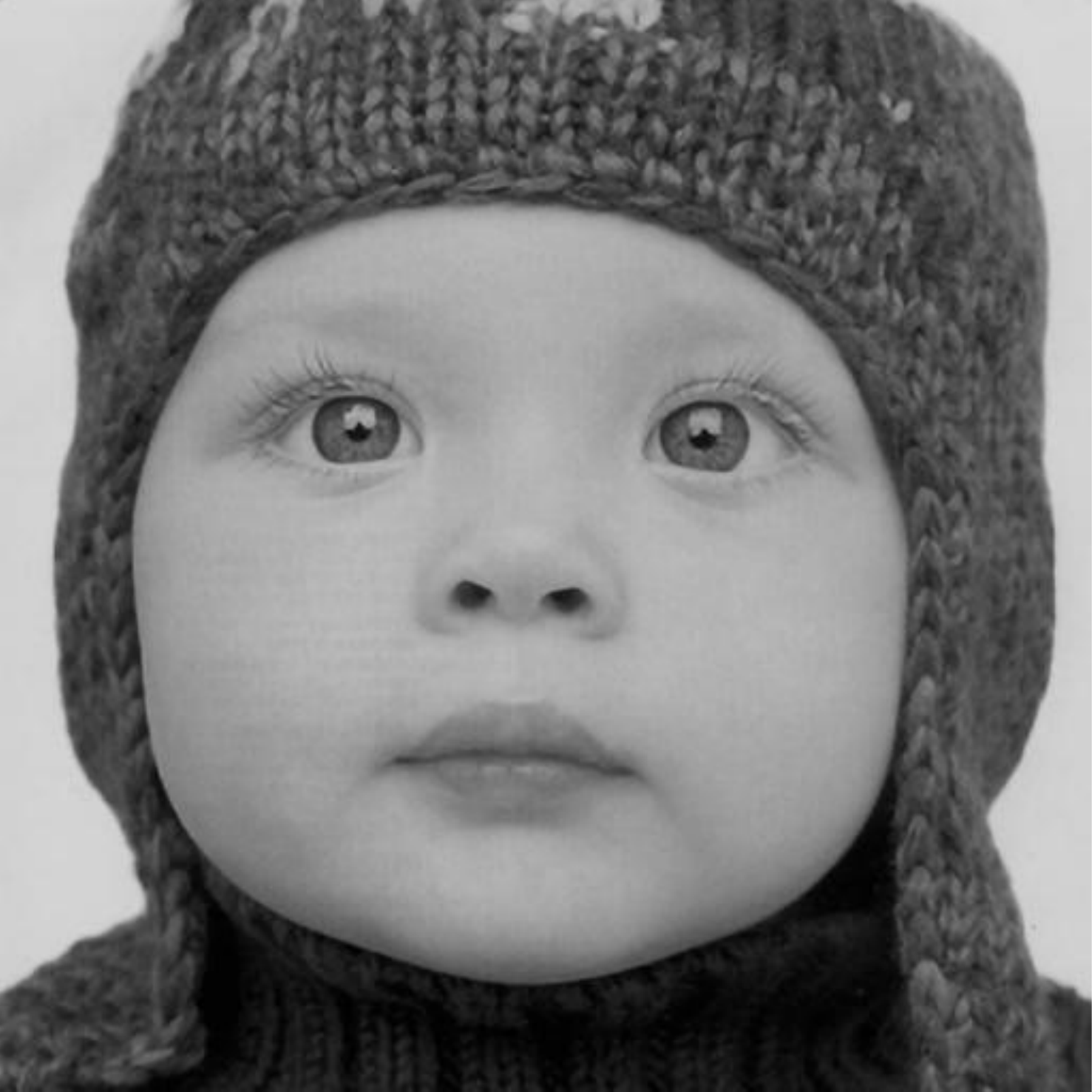} 
\end{subfigure}
\begin{subfigure}[b]{0.19\textwidth}
\includegraphics[width=0.95\textwidth]{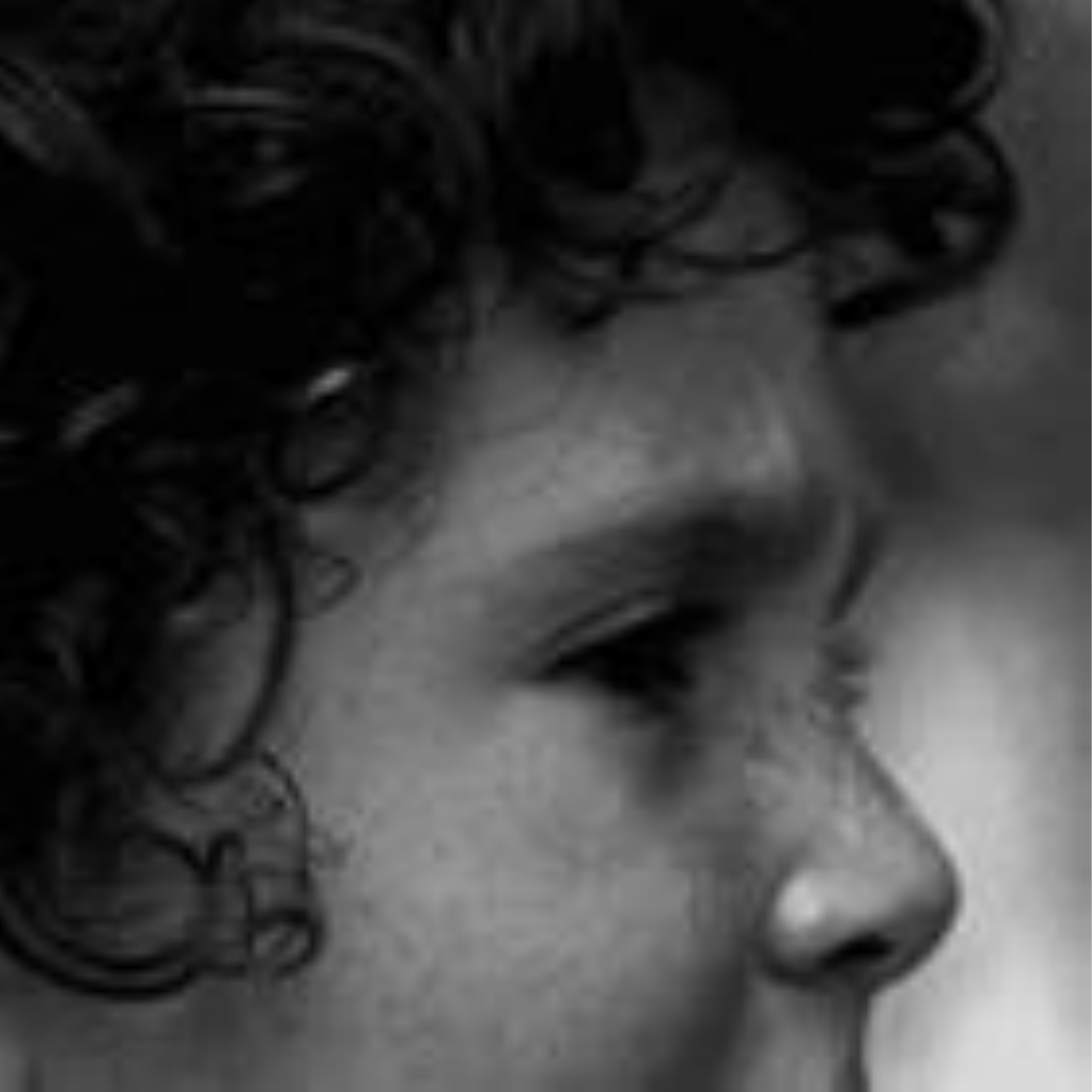} 
\caption{LR}
\end{subfigure}
\begin{subfigure}[b]{0.19\textwidth}
\includegraphics[width=0.95\textwidth]{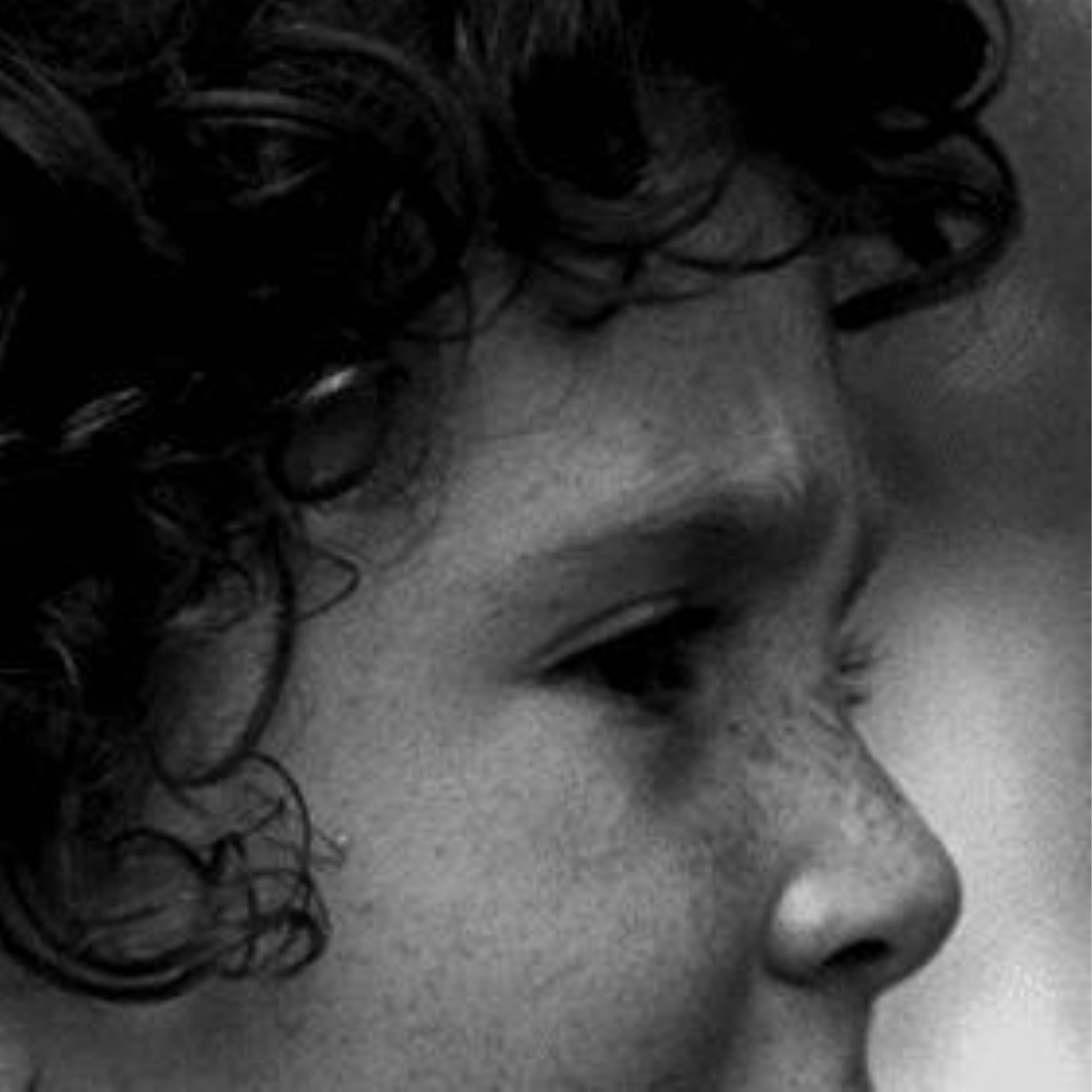} 
\caption{HR}
\end{subfigure}
\begin{subfigure}[b]{0.19\textwidth}
\includegraphics[width=0.95\textwidth]{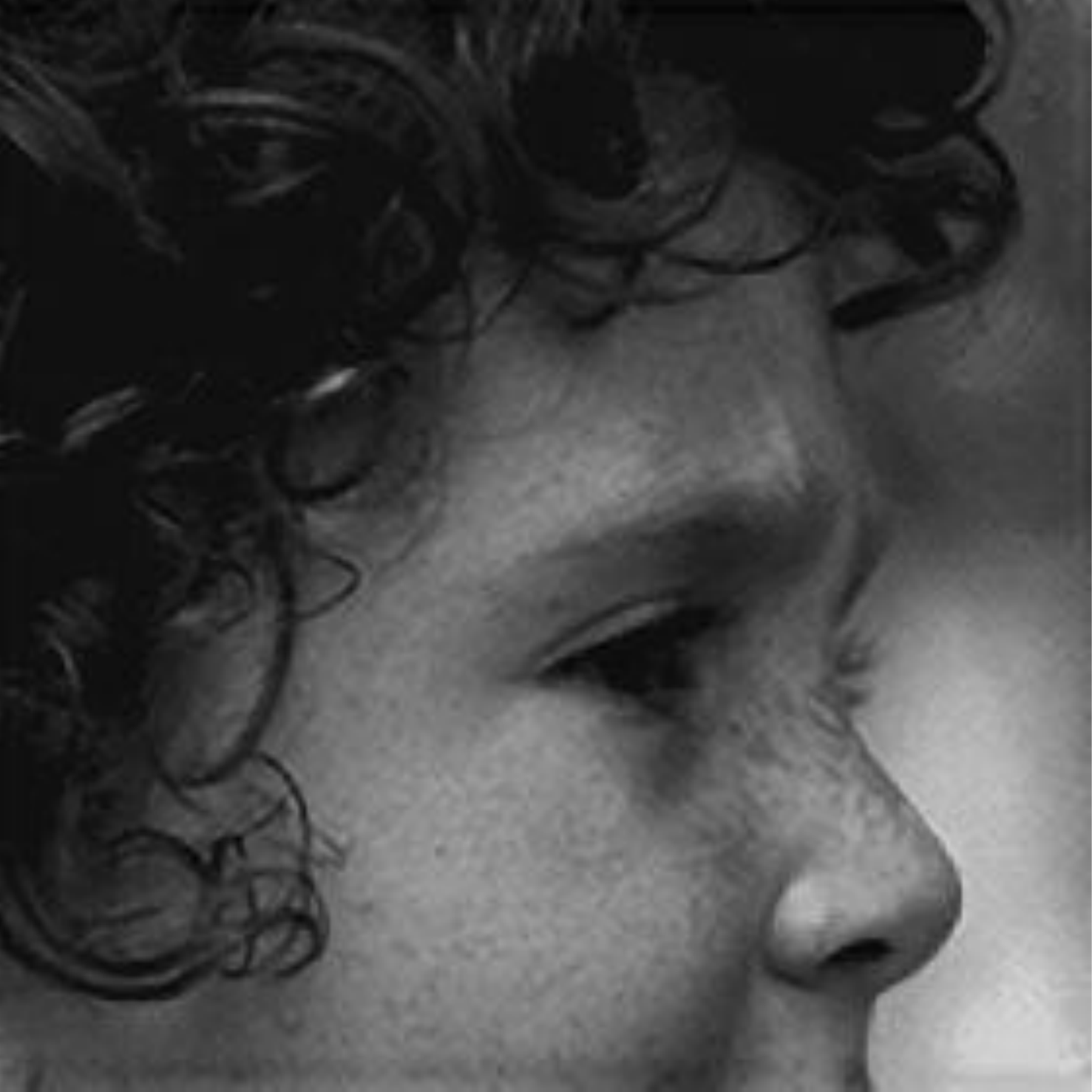} 
\caption{WaveSR}
\end{subfigure}
\begin{subfigure}[b]{0.19\textwidth}
\includegraphics[width=0.95\textwidth]{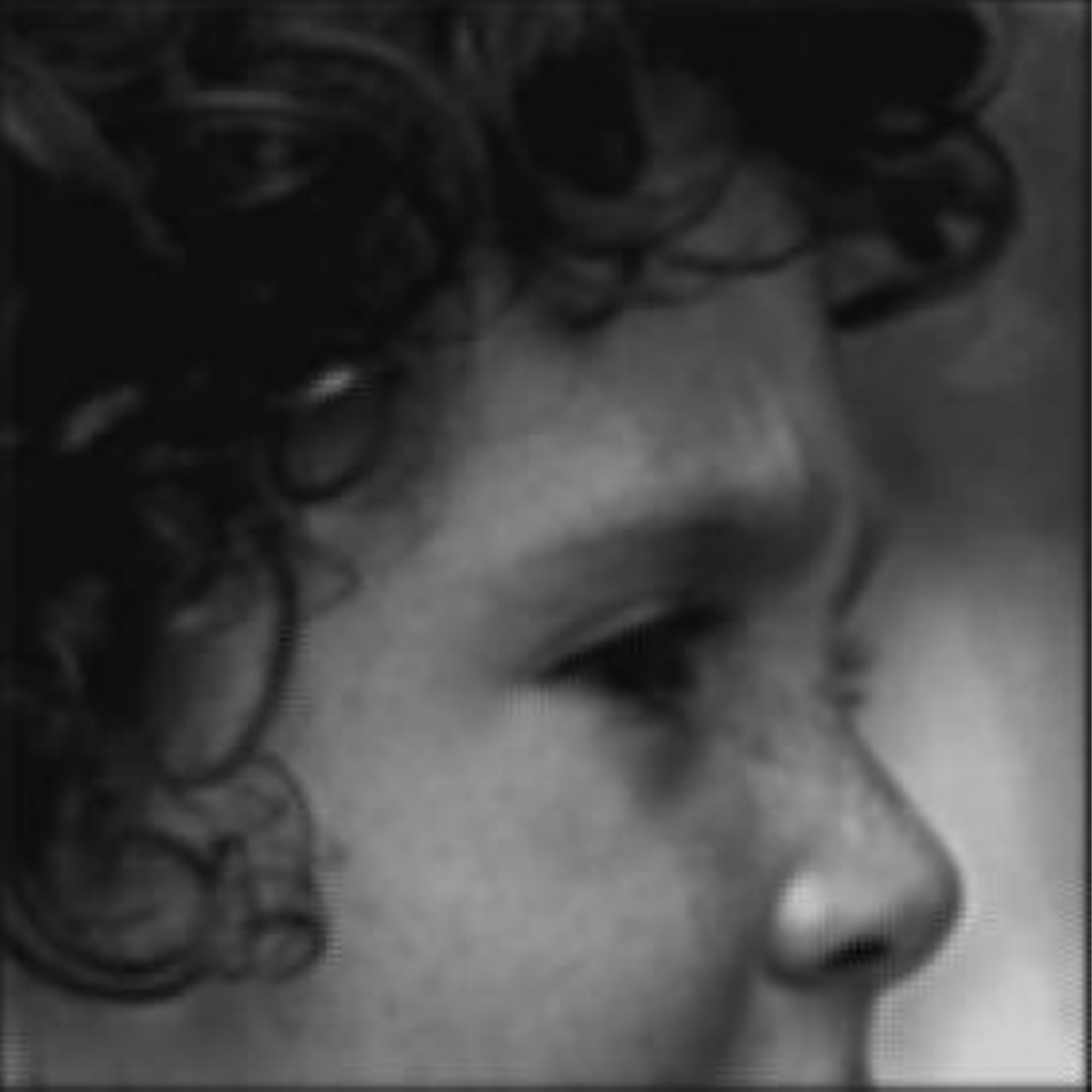} 
\caption{WSR}
\end{subfigure}
\begin{subfigure}[b]{0.19\textwidth}
\includegraphics[width=0.95\textwidth]{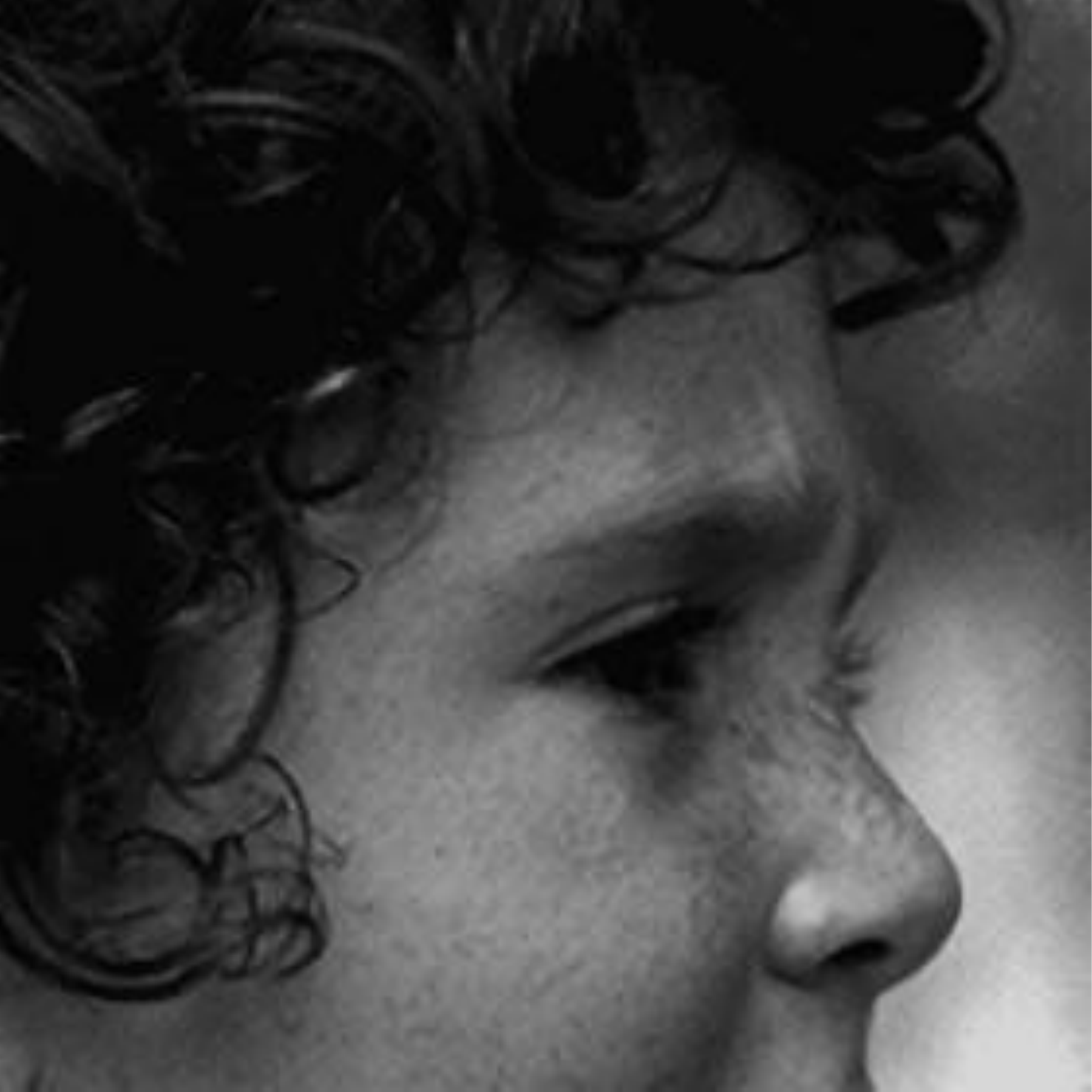} 
\caption{LPSR}
\end{subfigure}
{2$\times$ magnification} 
{\vskip 1.5mm}

\begin{subfigure}[b]{0.19\textwidth}
\includegraphics[width=0.95\textwidth]{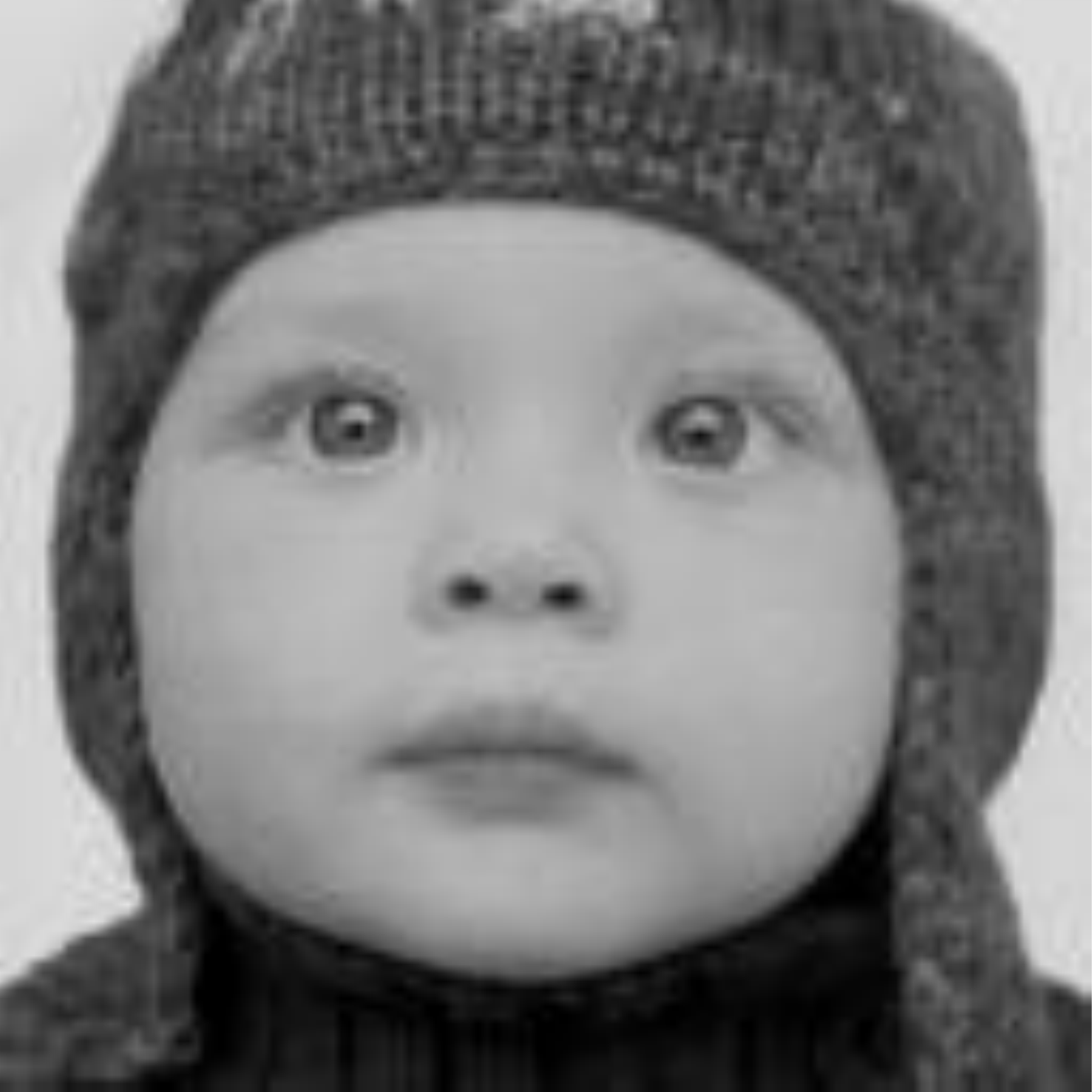} 
\end{subfigure}
\begin{subfigure}[b]{0.19\textwidth}
\includegraphics[width=0.95\textwidth]{./srnet_result/0_2mag_hr.pdf} 
\end{subfigure}
\begin{subfigure}[b]{0.19\textwidth}
\includegraphics[width=0.95\textwidth]{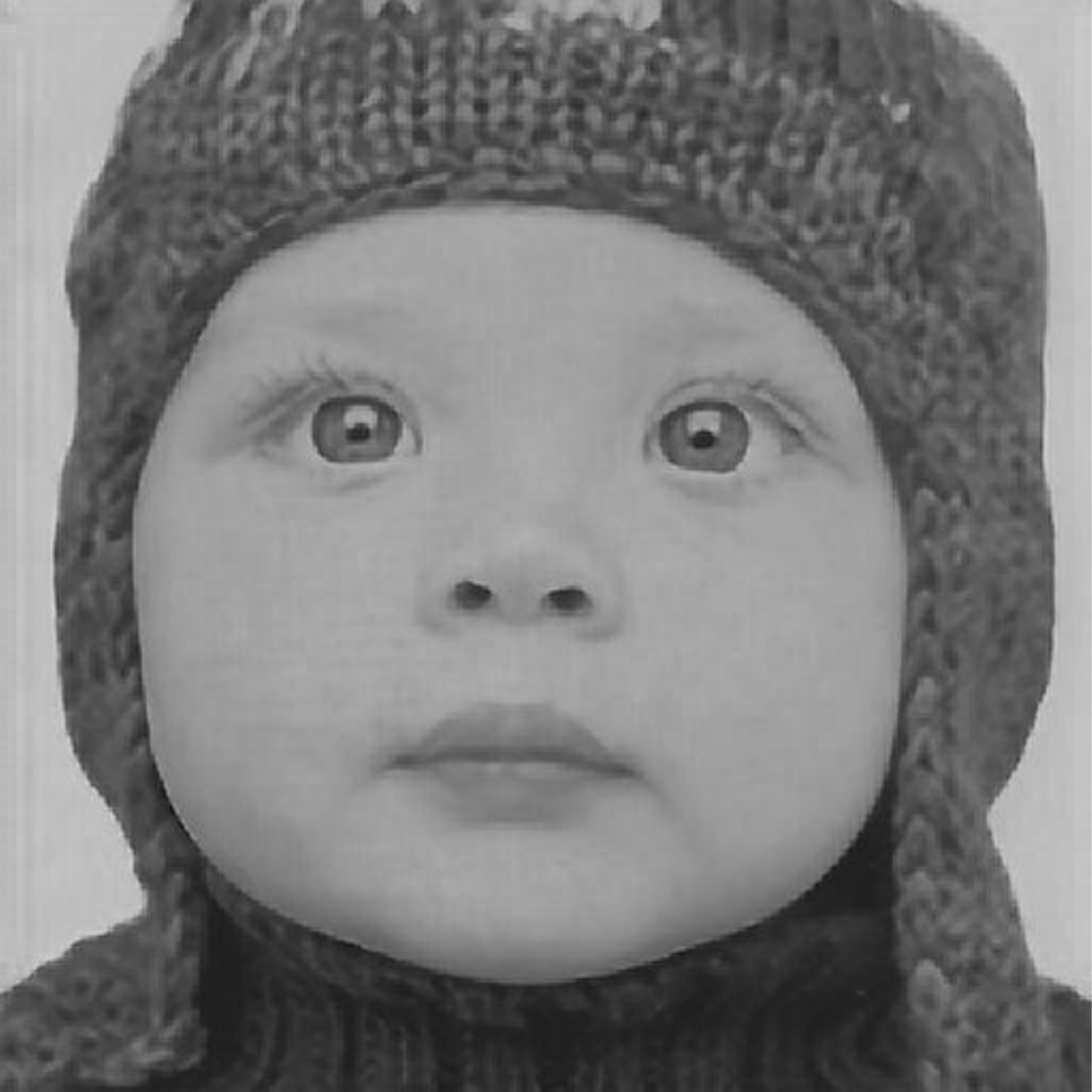} 
\end{subfigure}
\begin{subfigure}[b]{0.19\textwidth}
\includegraphics[width=0.95\textwidth]{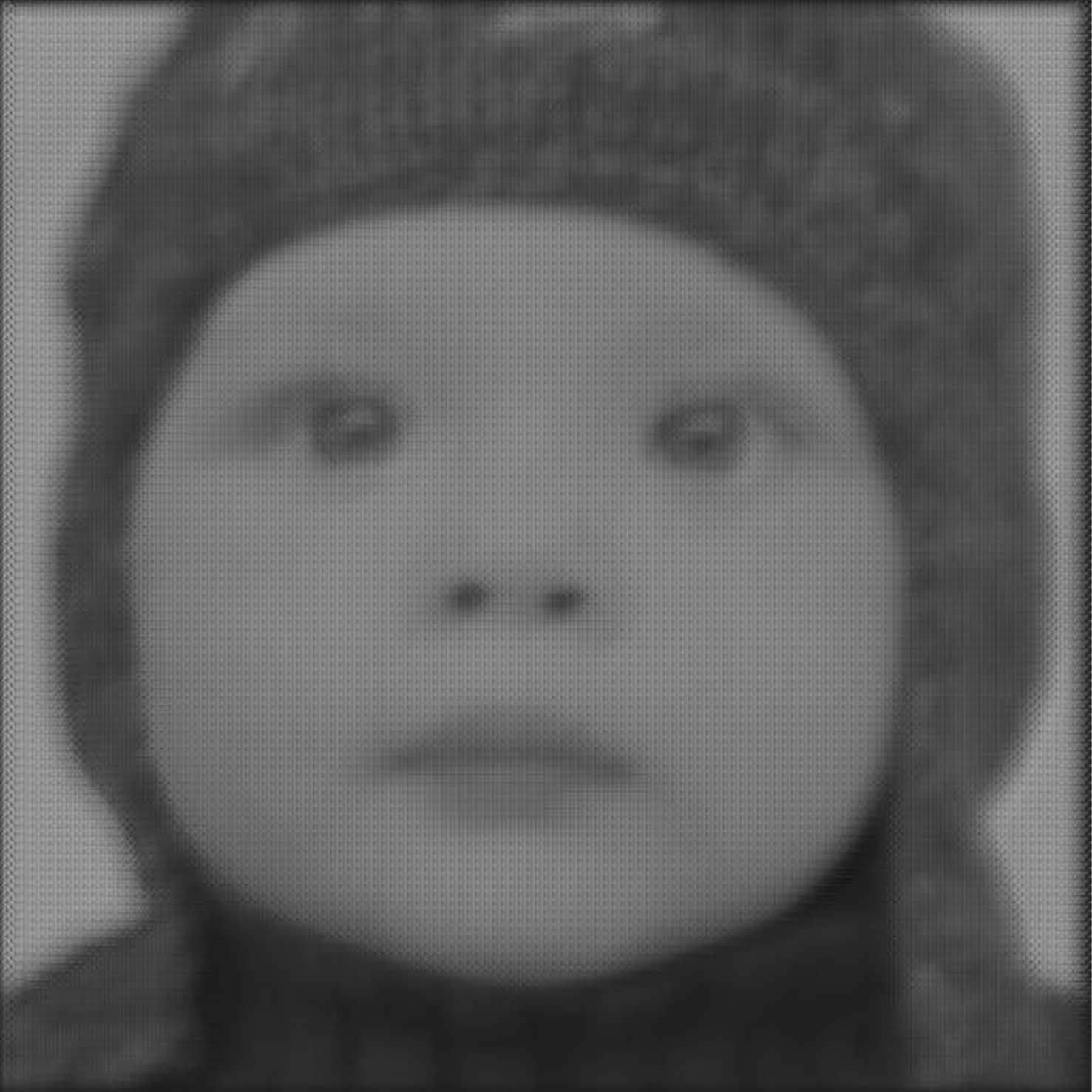} 
\end{subfigure} 
\begin{subfigure}[b]{0.19\textwidth}
\includegraphics[width=0.95\textwidth]{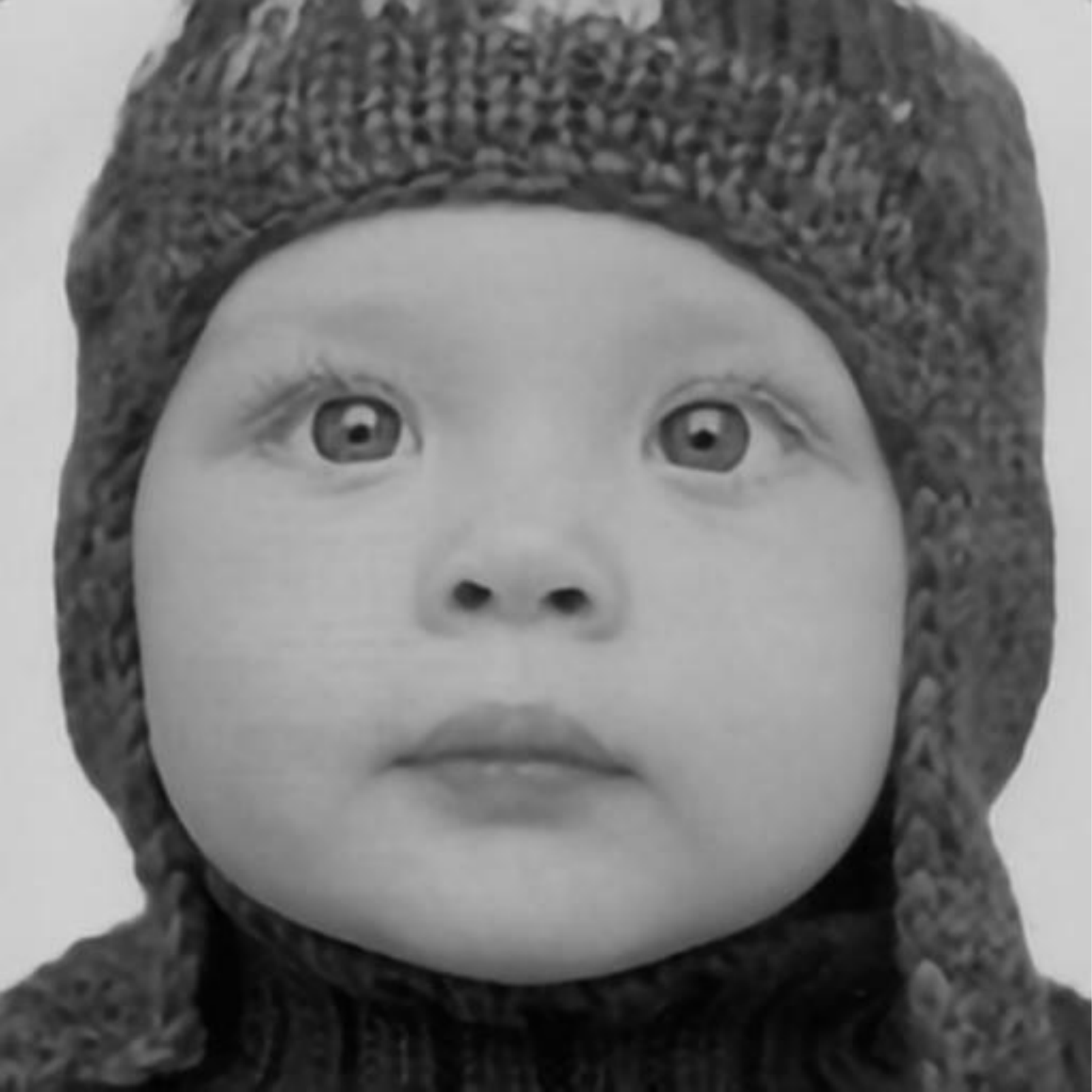} 
\end{subfigure}
\begin{subfigure}[b]{0.19\textwidth}
\includegraphics[width=0.95\textwidth]{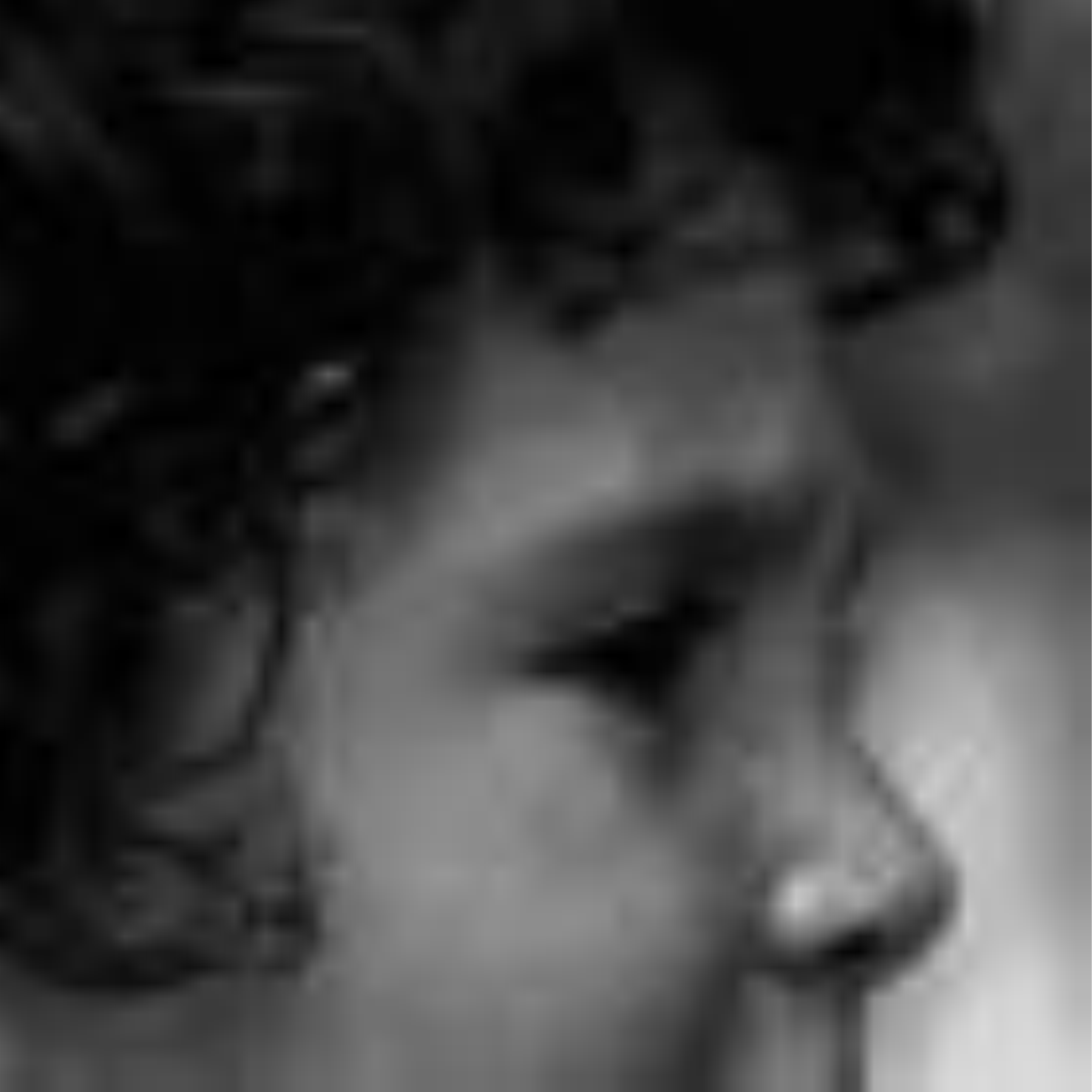} 
\caption{LR}
\end{subfigure}
\begin{subfigure}[b]{0.19\textwidth}
\includegraphics[width=0.95\textwidth]{./srnet_result/3_2mag_hr.pdf} 
\caption{HR}
\end{subfigure}
\begin{subfigure}[b]{0.19\textwidth}
\includegraphics[width=0.95\textwidth]{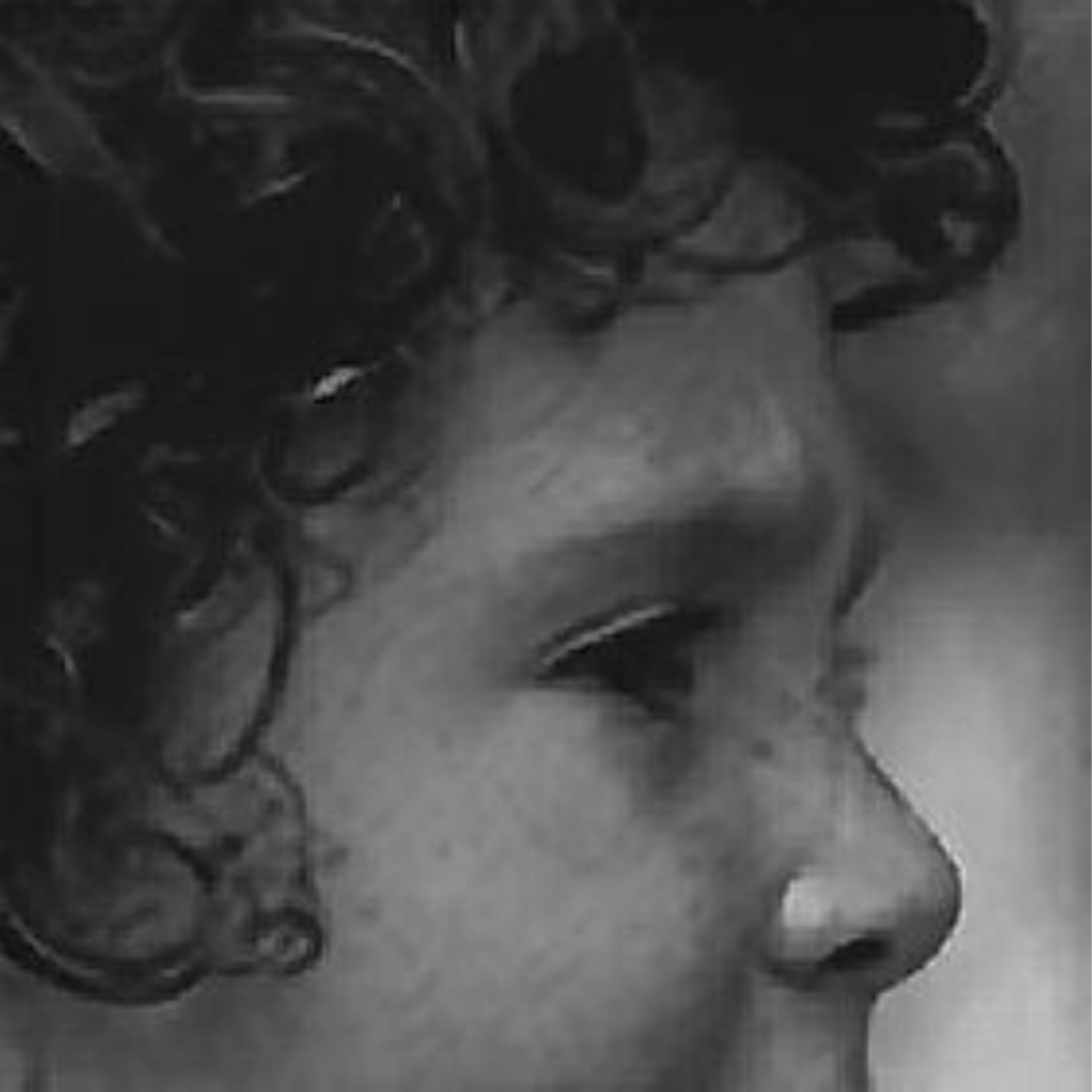} 
\caption{WaveSR}
\end{subfigure}
\begin{subfigure}[b]{0.19\textwidth}
\includegraphics[width=0.95\textwidth]{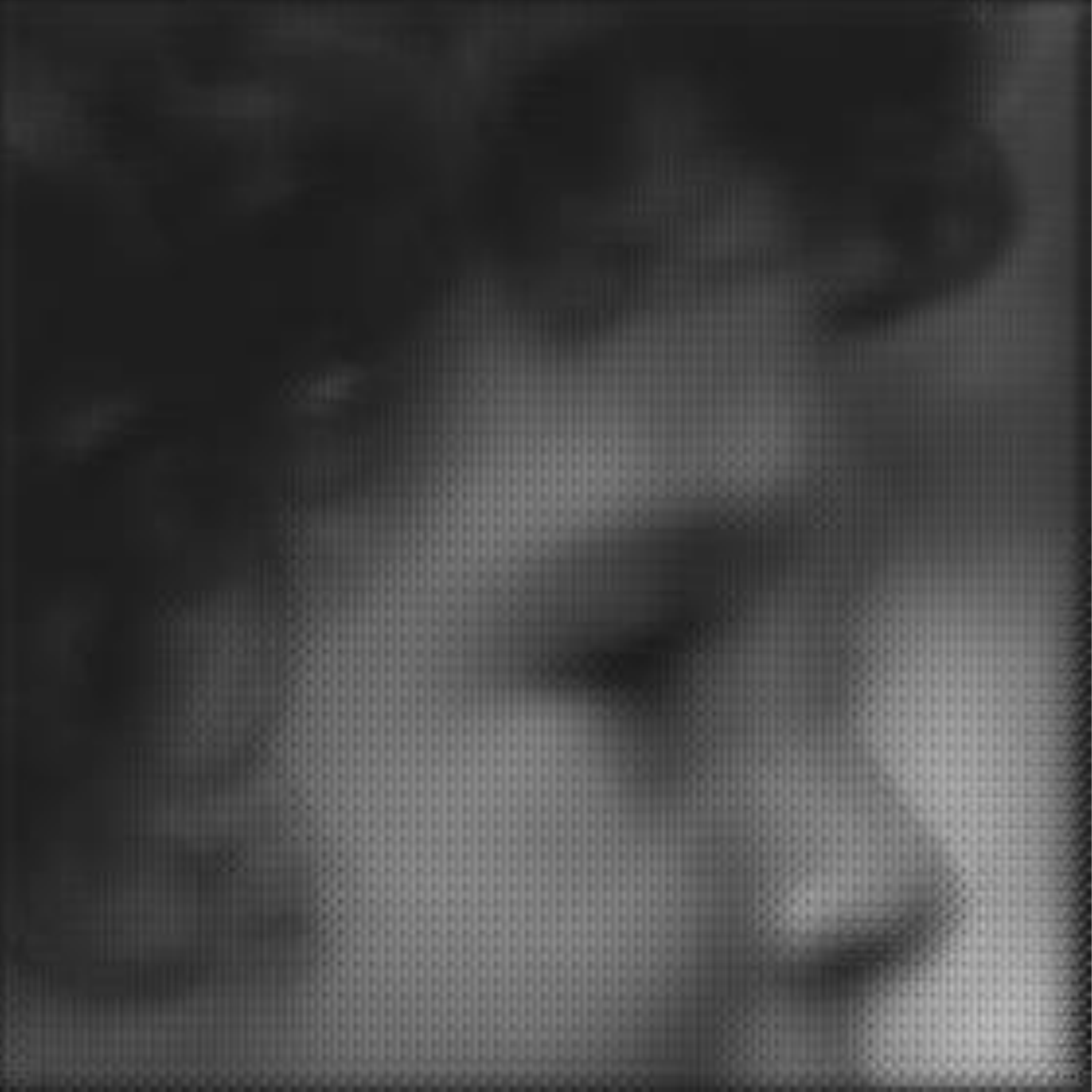} 
\caption{WSR}
\end{subfigure}
\begin{subfigure}[b]{0.19\textwidth}
\includegraphics[width=0.95\textwidth]{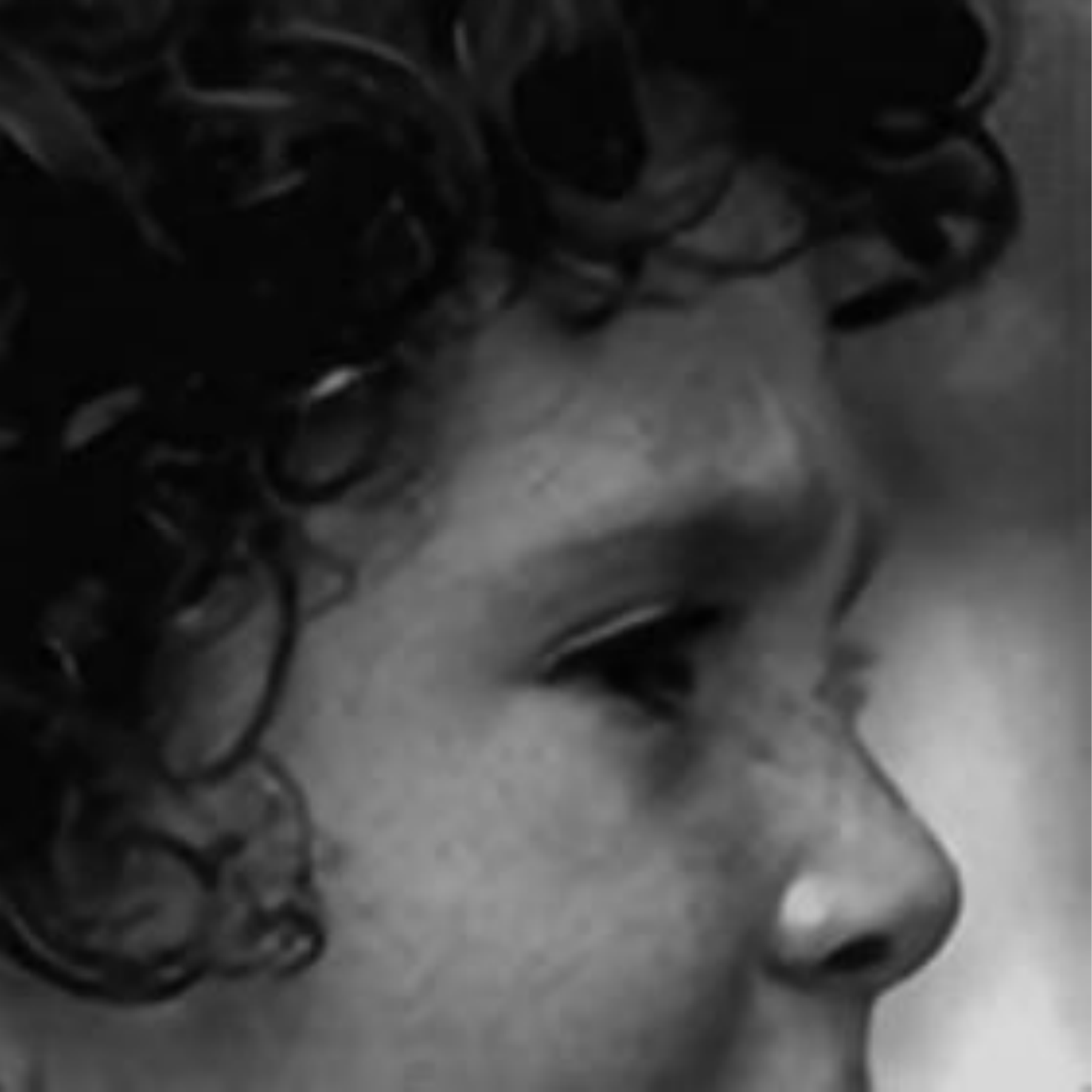} 
\caption{LPSR}
\end{subfigure}
{4$\times$ magnification}
{\vskip 1.5mm}

\begin{subfigure}[b]{0.19\textwidth}
\includegraphics[width=0.95\textwidth]{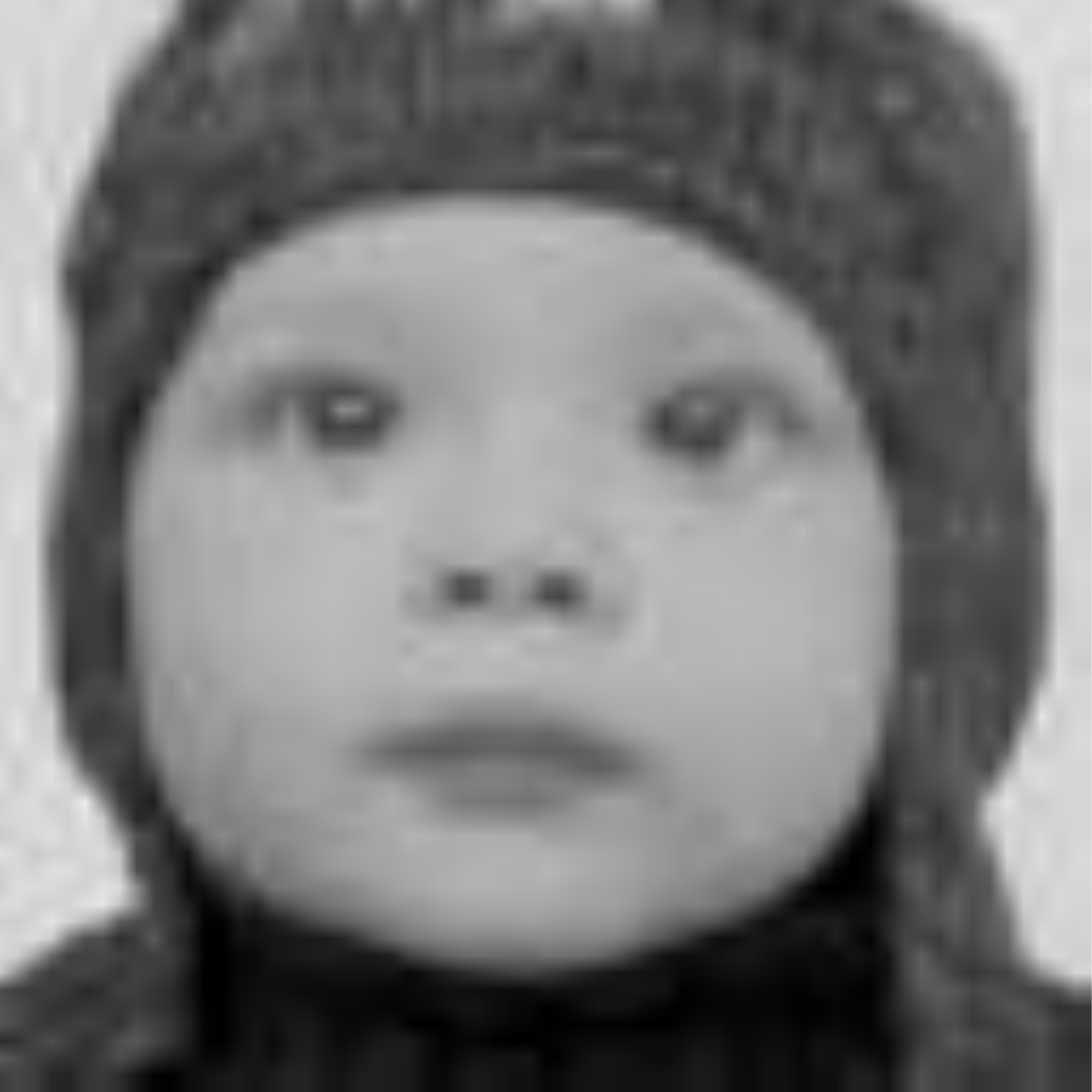} 
\end{subfigure}
\begin{subfigure}[b]{0.19\textwidth}
\includegraphics[width=0.95\textwidth]{./srnet_result/0_2mag_hr.pdf} 
\end{subfigure}
\begin{subfigure}[b]{0.19\textwidth}
\includegraphics[width=0.95\textwidth]{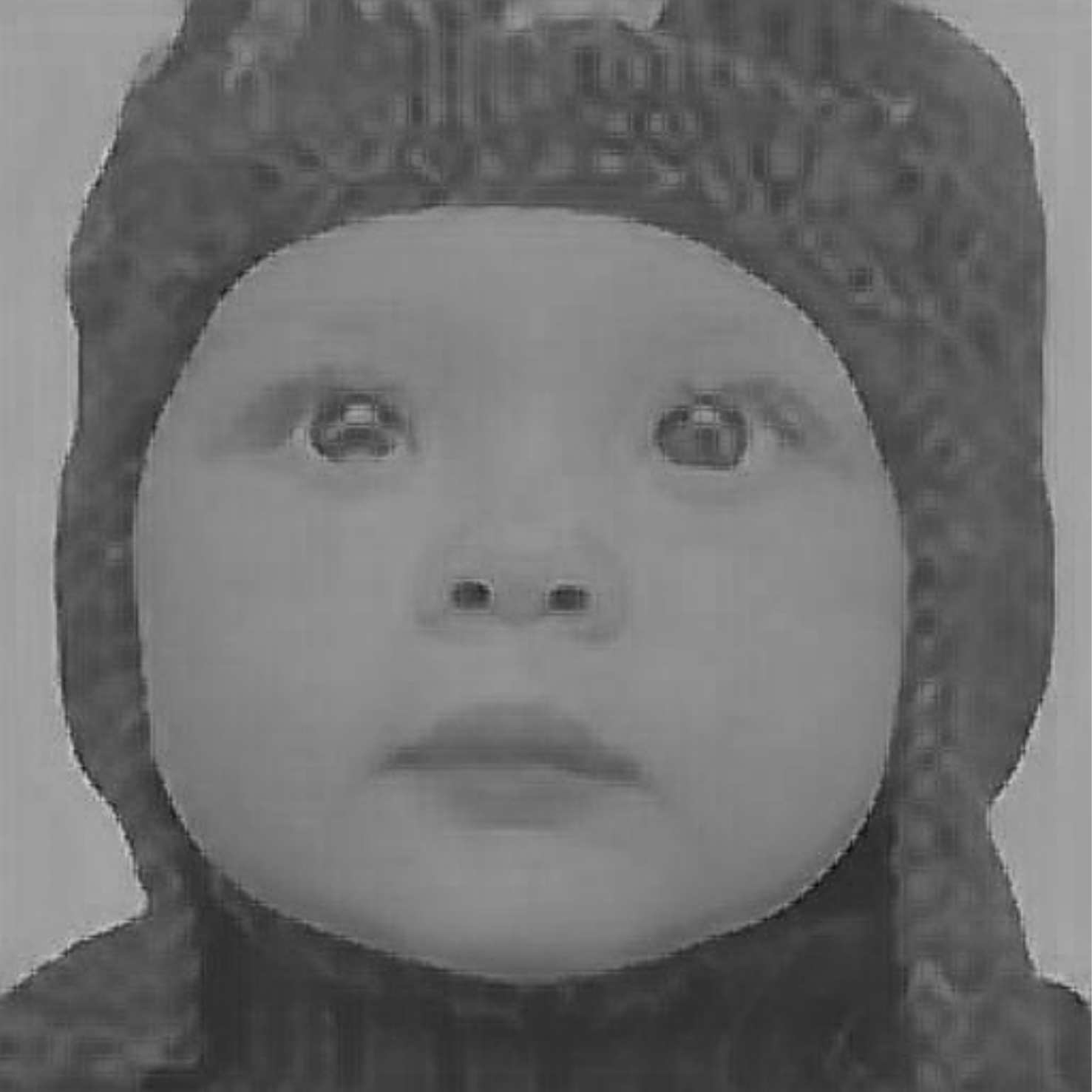} 
\end{subfigure}
\begin{subfigure}[b]{0.19\textwidth}
\includegraphics[width=0.95\textwidth]{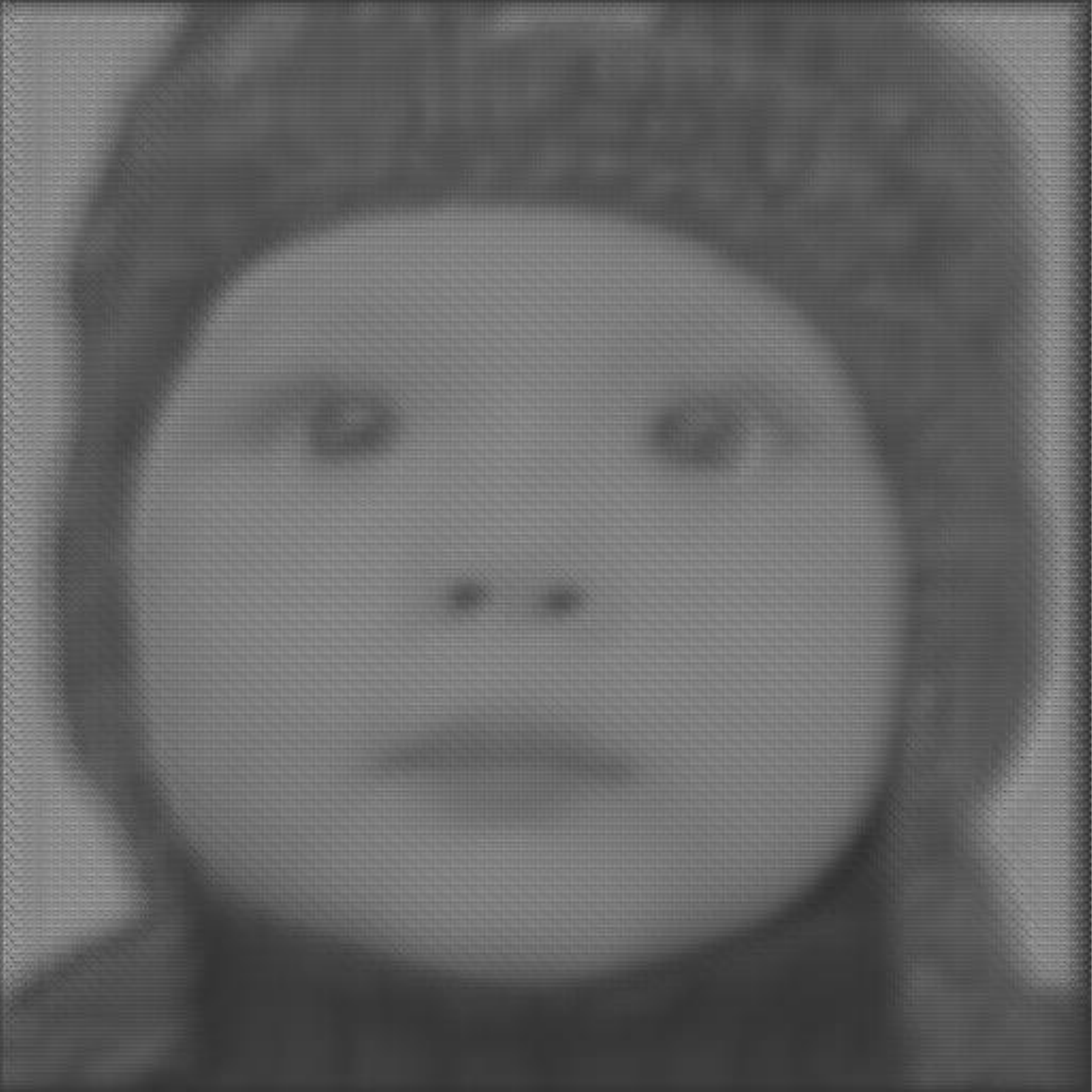} 
\end{subfigure} 
\begin{subfigure}[b]{0.19\textwidth}
\includegraphics[width=0.95\textwidth]{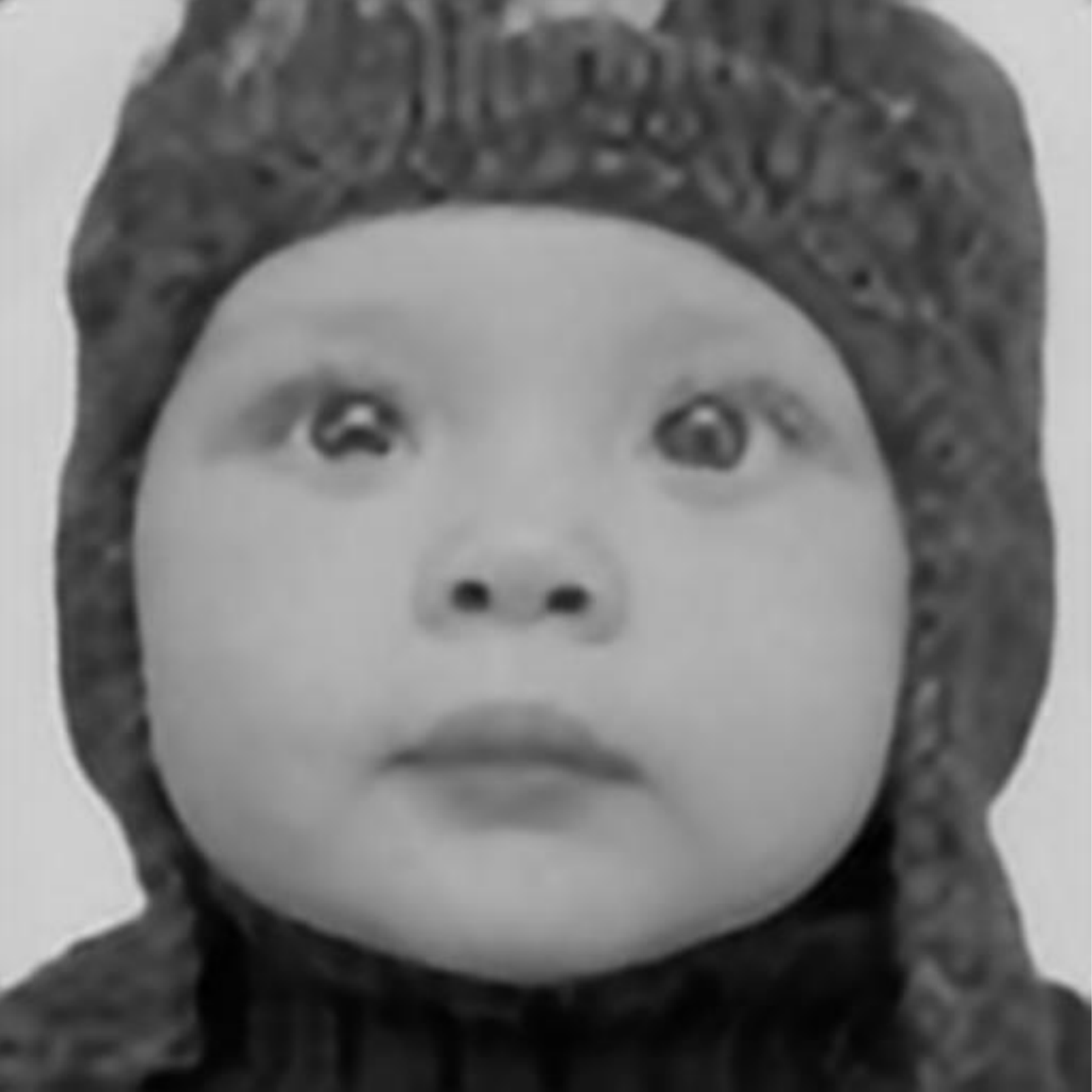} 
\end{subfigure}
\begin{subfigure}[b]{0.19\textwidth}
\includegraphics[width=0.95\textwidth]{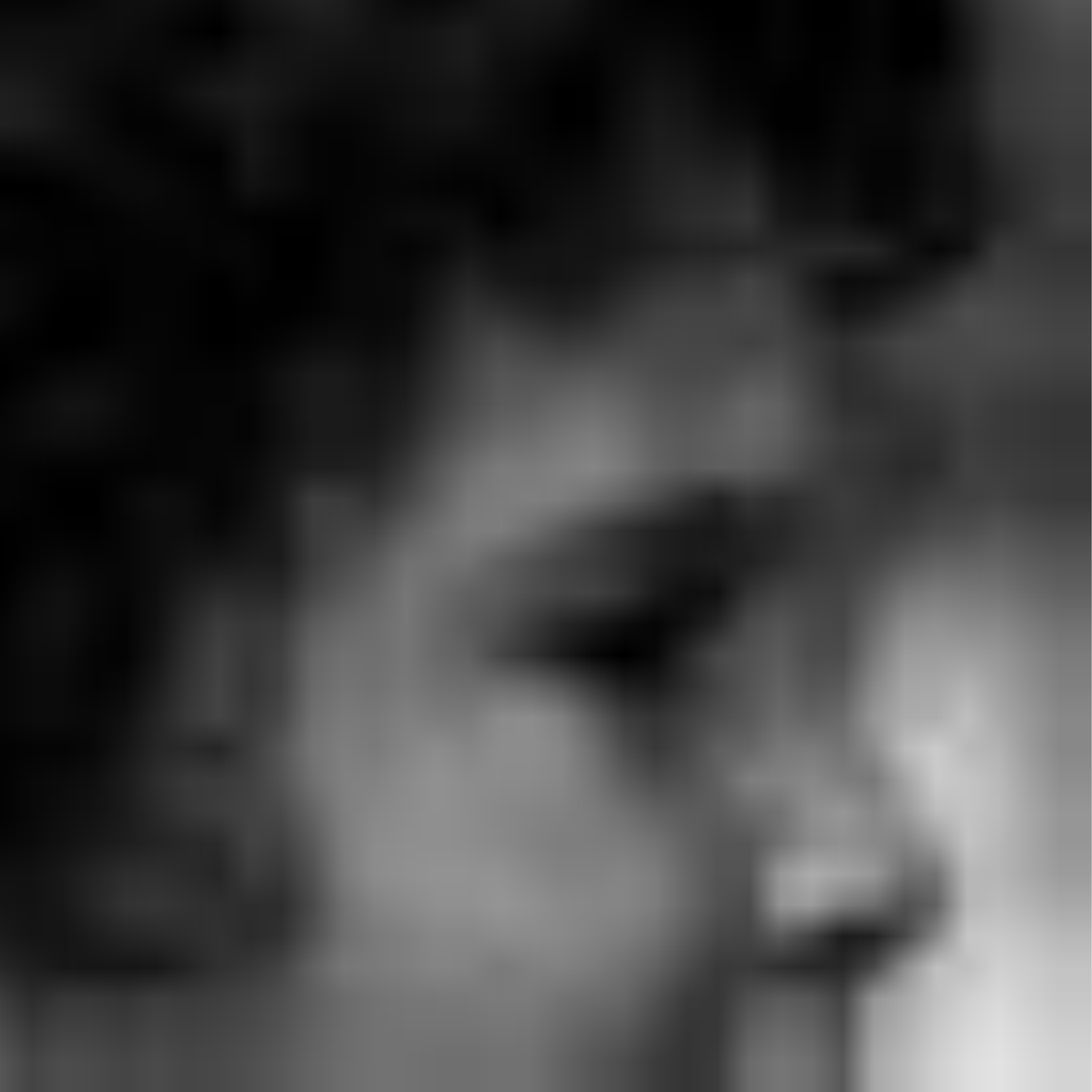} 
\caption{LR}
\end{subfigure}
\begin{subfigure}[b]{0.19\textwidth}
\includegraphics[width=0.95\textwidth]{./srnet_result/3_2mag_hr.pdf} 
\caption{HR}
\end{subfigure}
\begin{subfigure}[b]{0.19\textwidth}
\includegraphics[width=0.95\textwidth]{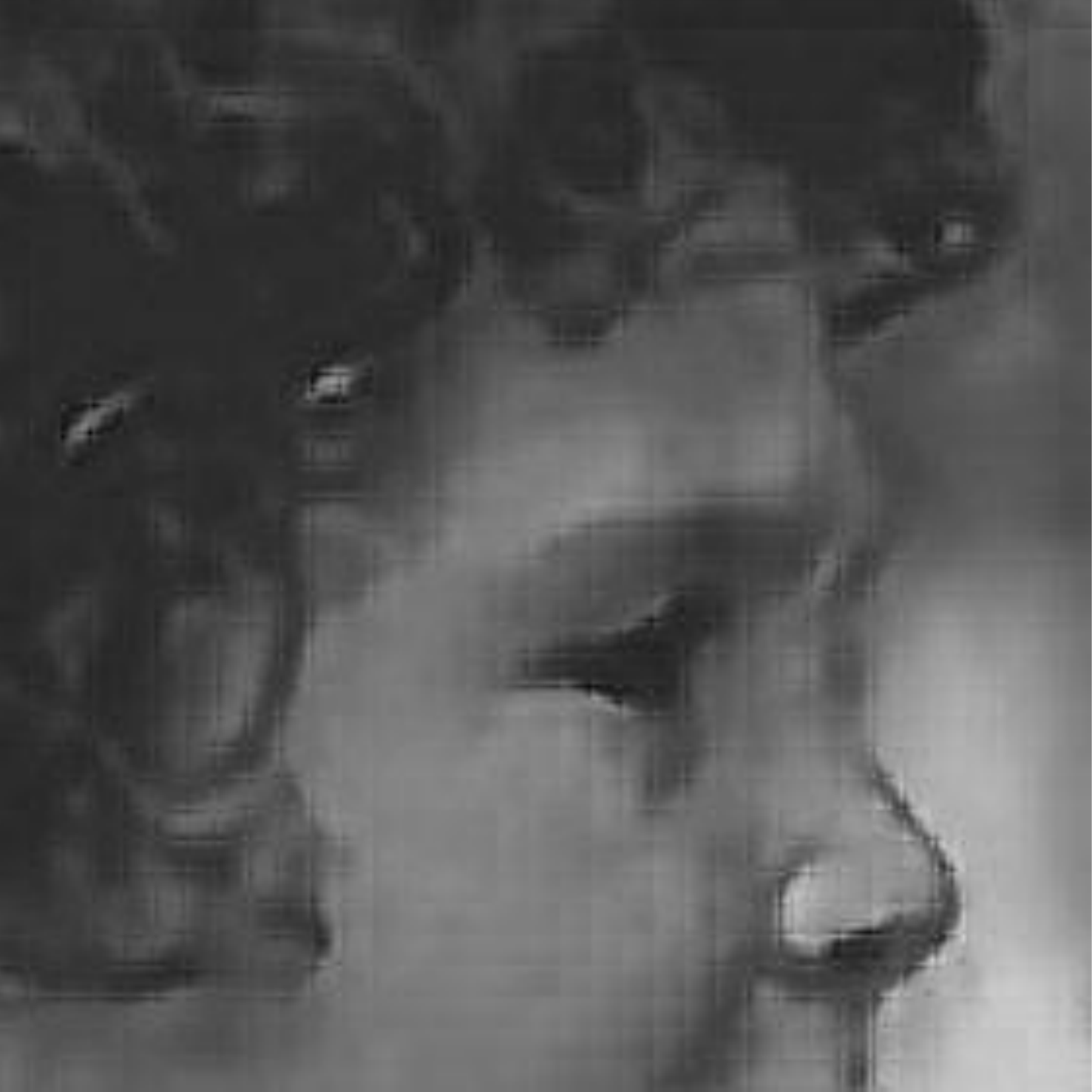} 
\caption{WaveSR}
\end{subfigure}
\begin{subfigure}[b]{0.19\textwidth}
\includegraphics[width=0.95\textwidth]{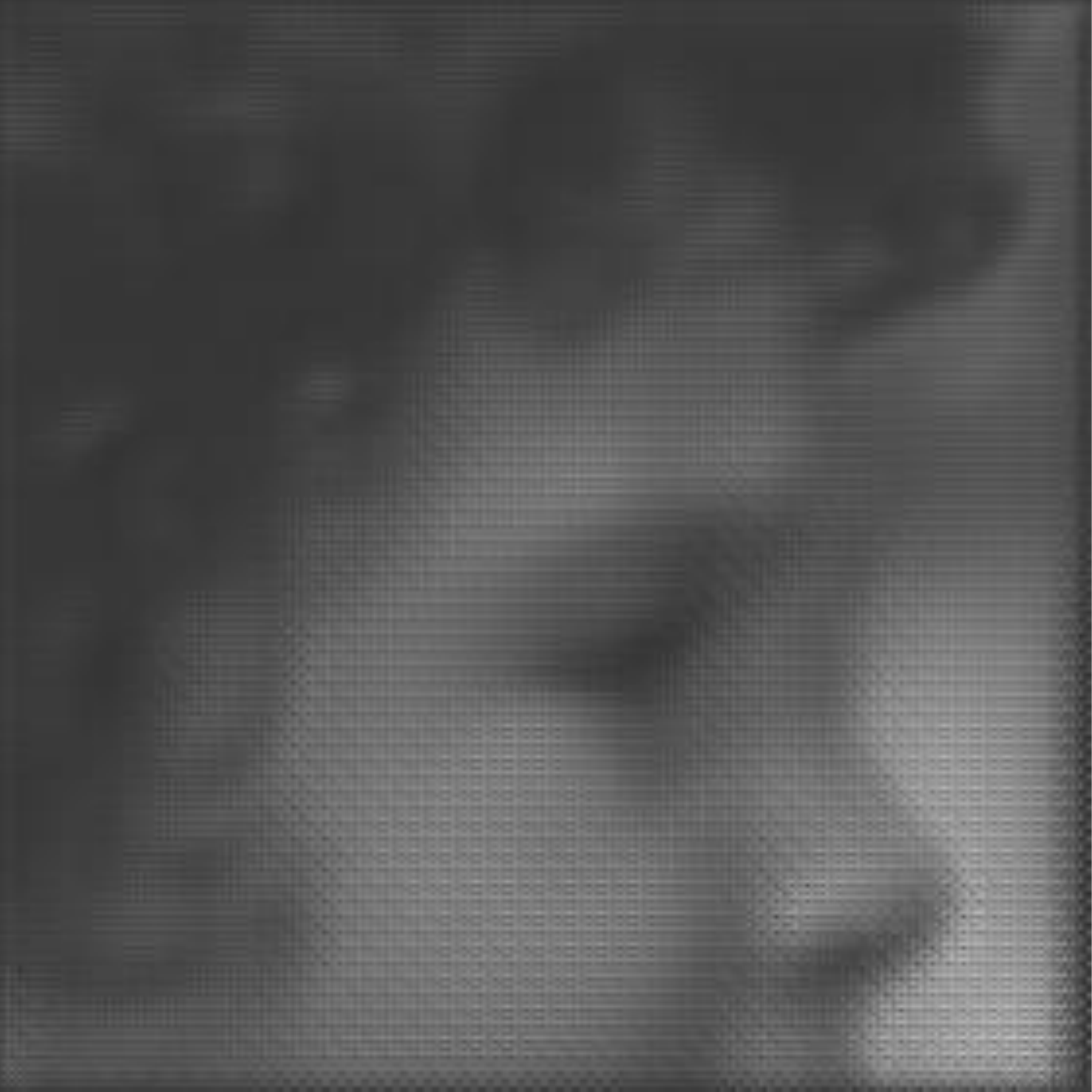} 
\caption{WSR}
\end{subfigure}
\begin{subfigure}[b]{0.19\textwidth}
\includegraphics[width=0.95\textwidth]{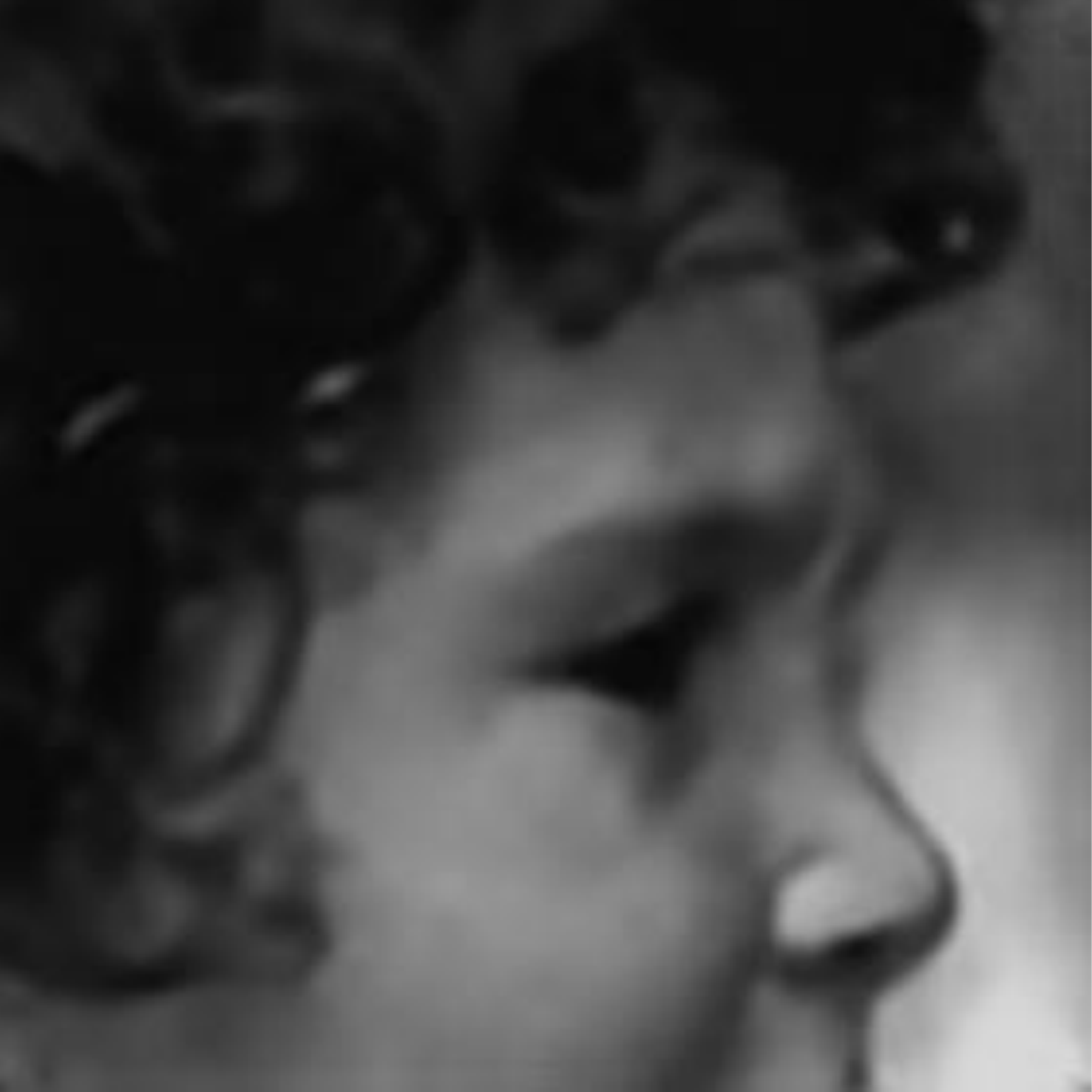} 
\caption{LPSR}
\end{subfigure}
{8$\times$ magnification}
\caption{Results of the super-resolution networks based on WaveletSRNet for Set5 dataset.}
\label{fig3}
\end{figure}

We evaluate super-resolution networks (WaveSR, WSR, LPSR) on two datasets (CelebA, Set5) using metrics PSNR and SSIM. For the result of WaveSR, we used our codes and could not reproduce results reported in \cite{WSR} despite the same options. The values in Table~\ref{t4} are obtained by applying our codes with the same environment to the three WaveletSRNet-based networks, namely, WaveSR, WSR, and LPSR. For CelebA, LPSR has top values among the three networks for all scales. In particular, for $\times2$ scale, there is the most significant gap when we change WPT to LPAE because PSNR is 36.04 dB for LPSR and 30.87 dB for WaveSR. This shows the power of LPAE reconstructing the image and having a learning method that fits the model's parameter on data distribution. If we focus on the comparison between WSR and LPSR, the PSNR value of WSR takes a sharp drop with increased scale. WSR for $\times2$ scale gets 31.93 dB, which is even higher than the PSNR of WaveSR, but for $\times4$ scale, WSR reaches 17.96 dB. This result describes the limit of WAE, which means that WAE does not consider multi-scale analysis and reconstruction of the image (c.f. Section~\ref{S:intro}). The same tendency appears once more in the results for the Set5 dataset. LPSR obtains the biggest PSNR, SSIM values among the three networks. Its PSNR values are 37.62 dB for $\times2$ scale, 32.25 dB for $\times4$ scale, and 27.07 dB for $\times8$ scale. For WSR, since Set5 is dissimilar to CelebA, which is the center-aligned data of constant size, reconstruction by WAE falls down to 25.70 dB in $\times2$ scale. For $\times4$ or $\times8$ scale, it is hard to do reconstruction using WSR.
Table.~\ref{t5} shows the comparison of our LPSR on Set5 with SOTA networks. Our PSNR values rank about the middle on average, which is good, considering the fact that LPSR is obtained by a simple change using LPAE from WaveSR.
Fig.~\ref{fig3} shows the reconstruction of WaveSR, WSR, LPSR for two images of Set5. WaveSR works well for$\times2$ scale, but as the scale is getting bigger, the reconstruction images of WaveSR are blurred and have a checkers pattern. However, LPSR gets high-quality reconstruction images for all scales.

\section{Conclusion and Future Works}
We organize LPAE, which assists in various problems from network acceleration to super-resolution. It reflects the structure of LP and consists of the encoder and the decoder. Using the encoder, we decompose an image into the approximation image with a low-resolution/frequency channel and the detail image with a high-frequency channel. The decoder recreates the original image correctly using the approximation and the detail. Three types of loss (approximation loss, sparsity loss, and reconstruction loss) enable us to obtain clear decomposed images and to achieve better reconstruction. Experiments in this paper show that LPAE makes the existing classification network light and preserves the original accuracy in classification. For super-resolution, it accomplishes better performance than the established wavelet-based model. In the future, we plan to explore a range of applications to different problems such as generative models, image compression, and character recognition.

\captionsetup[subfigure]{labelformat=empty}

\section{Acknowledgments}
$\dagger$This work was supported by Institute of Information \& communications Technology Planning \& Evaluation (IITP) grant funded by the Korea government(MSIT) (No.2021-0-00023, Developing a lightweight Korean text detection and recognition technology for complex disaster situations).

\noindent $\ddagger$This work was supported in part by National Research Foundation of Korea (NRF) [Grant Numbers 2015R1A5A1009350 and 2021R1A2C1007598], and by the `Ministry of Science and ICT' and NIPA via ``HPC Support'' Project.

%
%

\end{document}